\documentclass{article}

\usepackage{PRIMEarxiv}

\usepackage[utf8]{inputenc} 
\usepackage[T1]{fontenc}    
\usepackage{amsmath,amssymb,amsfonts}
\usepackage{algorithmic}
\usepackage{subfig}
\usepackage{graphicx}
\usepackage{textcomp}
\usepackage{tabularx}
\usepackage{booktabs}
\usepackage{url}
\usepackage{comment}
\usepackage{hyperref}
\usepackage{adjustbox}
\usepackage{eurosym}

\title{Autonomous Robotic Pruning in Orchards and Vineyards: a Review
\thanks{This work has been developed with the contribution of the Politecnico di Torino Interdepartmental Centre for Service Robotics (PIC4SeR). \url{https://pic4ser.polito.it/}} 
}

\author{
  Alessandro Navone\\
  Department of Electronics and Telecommunications, \\
  Politecnico di Torino \\
  Torino, Italy\\
  \texttt{alessandro.navone@polito.it} \\
   \And
  Mauro Martini\\
  Department of Electronics and Telecommunications, \\
  Politecnico di Torino \\
  Torino, Italy\\
  \texttt{mauro.martini@polito.it} \\
   \And
    Marcello Chiaberge\\
  Department of Electronics and Telecommunications, \\
  Politecnico di Torino \\
  Torino, Italy\\
  \texttt{marcello.chiaberge@polito.it} \\
}

\begin{document}
\maketitle

\begin{abstract}
Manual pruning is labor intensive and represents up to $25\%$ of annual labor costs in fruit production, notably in apple orchards and vineyards where operational challenges and cost constraints limit the adoption of large-scale machinery. In response, a growing body of research is investigating compact, flexible robotic platforms capable of precise pruning in varied terrains, particularly where traditional mechanization falls short.

This paper reviews recent advances in autonomous robotic pruning for orchards and vineyards, addressing a critical need in precision agriculture. Our review examines literature published between 2014 and 2024, focusing on innovative contributions across key system components. Special attention is given to recent developments in machine vision, perception, plant skeletonization, and control strategies, areas that have experienced significant influence from advancements in artificial intelligence and machine learning. The analysis situates these technological trends within broader agricultural challenges, including rising labor costs, a decline in the number of young farmers, and the diverse pruning requirements of different fruit species such as apple, grapevine, and cherry trees.

By comparing various robotic architectures and methodologies, this survey not only highlights the progress made toward autonomous pruning but also identifies critical open challenges and future research directions. The findings underscore the potential of robotic systems to bridge the gap between manual and mechanized operations, paving the way for more efficient, sustainable, and precise agricultural practices.
\end{abstract}

\keywords{Pruning \and Autonomous Robots \and Review \and Orchards \and Vineyards.}

\section{Introduction}
Every year, fruit orchards need intensive labor operations such as harvesting and pruning, which are still conducted manually by the majority of farmers. Pruning is a repetitive and costly process that aims to ensure the growth of new plant nodes after winter seasons.
For instance, recent studies show that the pruning operation covers from 20$\%$ to 25$\%$ of the overall annual labor cost for an apple farmer~\cite{galinato2022costs}.

Fruit production is a key agricultural industry in both the European Union (EU) and the United States (US). In 2022, fruit harvesting in the EU was valued at €27.3 billion, with Spain, Italy, and Poland leading in production. The EU produced 14.7 million tonnes of pome fruit, 10.5 million tonnes of citrus fruit, 6.3 million tonnes of stone fruit, 2.6 million tonnes of tropical fruit, 1.1 million tonnes of nuts, and 0.7 million tonnes of berries~\cite{europe2024fruit}. In the same year, the US recorded $\$34.2$ billion in sales from fruit, tree nuts, and berries, representing 6.3\% of total agricultural sales, according to the US Department of Agriculture (USDA) Economic Research Service. Moreover, vines covered 3.2 million hectares (ha) in the EU in 2020, with 2.2 million vineyard holdings. More importantly, $83.3\%$ had less than 1 ha of vineyards, confirming the prevalence of small activities and fields.

Despite the sector’s economic significance, the EU faces demographic and labor-related challenges. The proportion of young farmers has sharply declined, with only 6.5\% of EU farm managers under the age of 35 in 2020~\cite{europe2024farmers}. Additionally, the availability of manual labor has decreased in recent years. From 2010 to 2019, labor costs—typically around 30\% of total production costs for fruit and nut trees—have increased for several reasons. A recent USDA Economic Research Service report discusses how producers are responding to these rising costs and labor shortages~\cite{calvin2022adjusting}. Long-term strategies include the use of temporary worker visas such as the H-2A program and, in some cases, a reduction in domestic production.

Automatic machinery has emerged as a possible support for boosting efficiency in agriculture and reducing costs without requiring intensive manual operations. 
Several recent studies discussed the feasibility of adopting a mechanized solution for pruning in both vineyards and orchards~\cite{allegro2022effects, galinato2022costs}. Agriculture experts are advancing methods to organize canopies in walls or trellis structures to foster mechanized pruning~\cite{he_sensing_2018}.
However, huge machines have limitations. First, they cannot be applied on slopes and hills, typical of European vineyards and orchards. Moreover, the investment required is not ignorable and affordable for small and medium-sized producers, ranging around \text{\euro}$12k$ for the machine itself to be coupled with a tractor, without considering maintenance and insurance costs. Finally, a high-quality economy, such as a grapevine, requires precise and careful operations. Often, custom pruning strategies are adopted by different wine producers.

\begin{figure}[!t]
    \centering
    \includegraphics[width=1\linewidth]{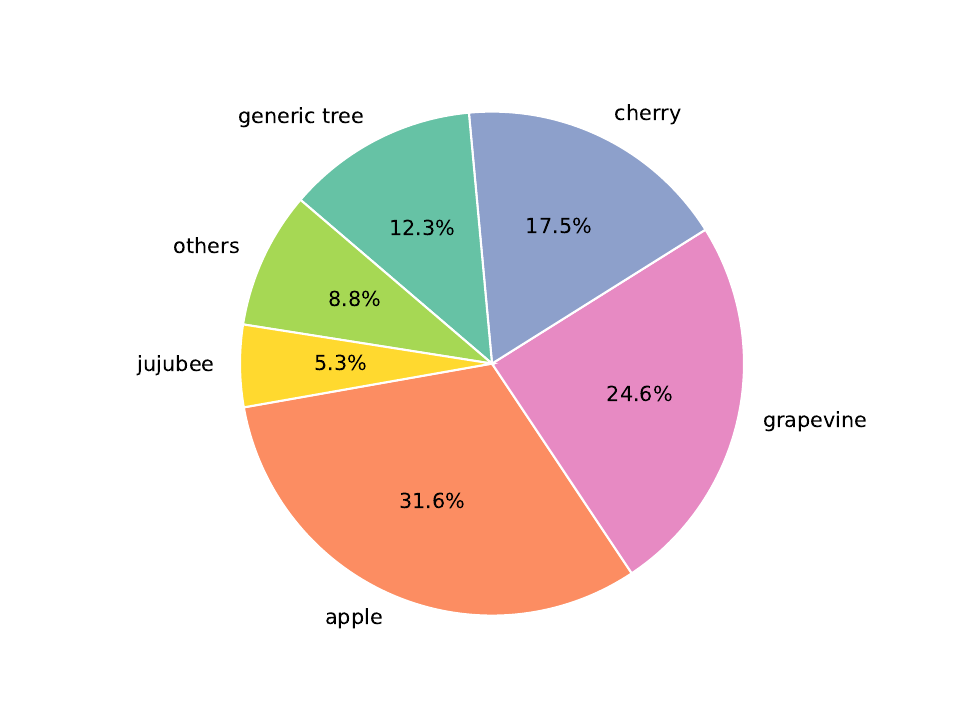}
    \caption{Pie chart of the distribution of the addressed crops in the analyzed papers.}
    \label{fig:pie-chart}
\end{figure}

Apples and vines represent the principal production industry, and, hence, the main focus of research advancement in the direction of technological solutions. Indeed, among papers published from 2014 to 2024, as presented in Fig.~\ref{fig:pie-chart}, 31.6$\%$ of the studies consider the pruning of apple trees and 24.6$\%$ consider grapevine as application scenarios. A relevant part of the considered studies covers the pruning of cherry trees (17.5$\%$), while many works target generic trees, preferring not to target a specific fruit tree but addressing part of the methodology or analyzing the performance of a specific type of sensor. Finally, jujubees cover a smaller part of the analysis, while the rest of the considered work, which are grouped as "others", analyze sparsely other fruit trees such as walnut, avocado, pear, mango, and more.

Robotic pruning represents an optimal trade-off between manual and massively mechanized operations, leading to a flexible solution with small-medium size platforms.

A typical robotic platform for pruning is a complex system composed by different modules. In this review, we analyze the innovative contributions to each robotic component. Nonetheless, a special focus is devoted to machine vision, perception, plant skeletonization and reconstruction and control, considering the huge impact of recent Artificial Intelligence (AI) on these areas.

\subsection{Scope of the review}

\begin{figure}[!t]
    \centering
    \includegraphics[width=1\linewidth]{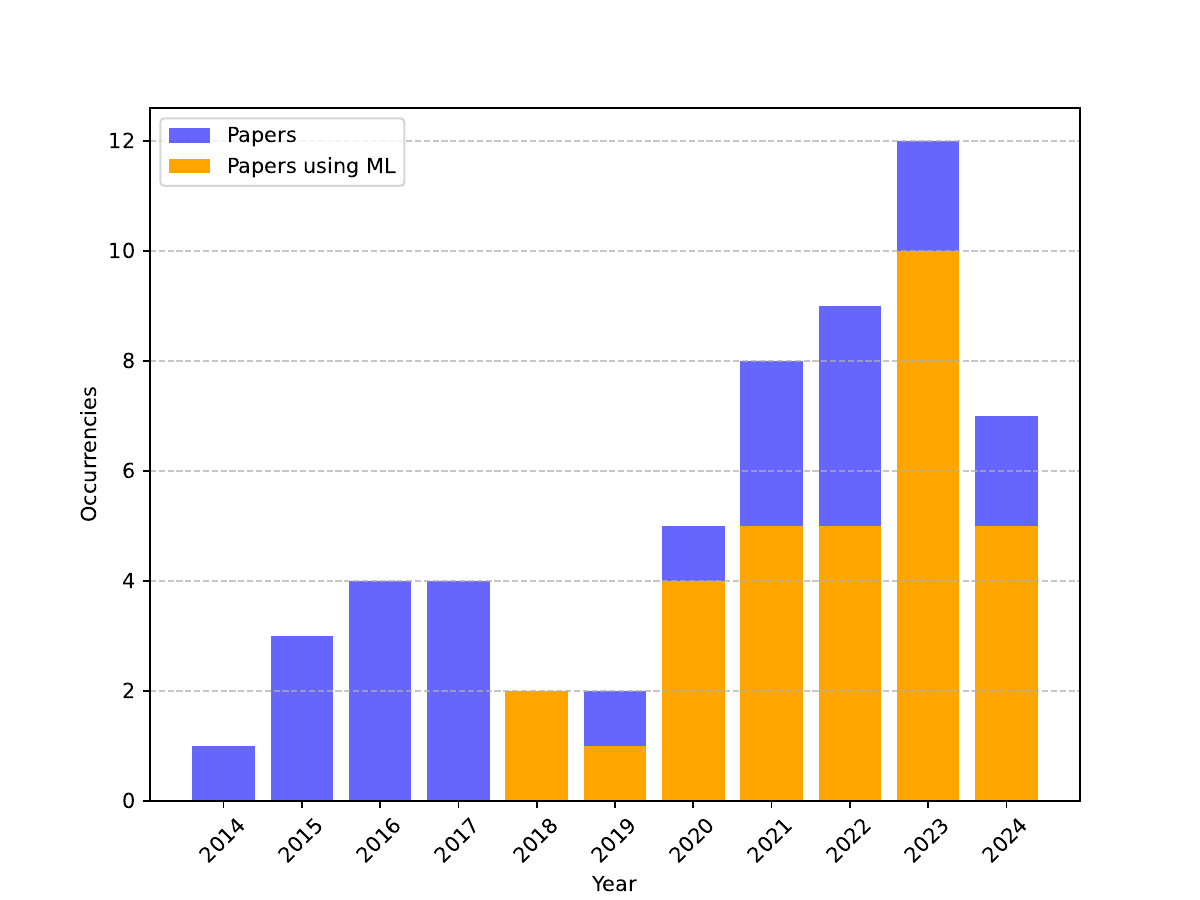}
    \caption{Distribution of the analyzed papers in the period of time from 2014 to 2024. The orange bars represent the paper using machine learning, showing a neat increase in recent years.}
    \label{fig:histo-years}
\end{figure}

Currently, autonomous robotic pruning in orchards and vineyards remains a niche area within the broader robotics literature focused on precision agriculture, as evidenced by the data presented in Fig.~\ref{fig:histo-years}. 
Only a limited number of review studies have been conducted in recent years, with the latest being published in 2021. This work aims to summarize the latest advancements in the field and outline potential future research directions. A first survey in the field of autonomous pruning was conducted by He and Schupp in 2018~\cite{he_sensing_2018}, in which sensing and perception techniques are explored solely in the context of apple orchards. It covers mainly the fields of tree training systems, pruning strategies, and plant structure reconstruction. Furthermore, the study outlined potential applications for automated pruning, suggesting a machine-friendly tree architecture and exemplifying pruning strategies that could streamline automated pruning decision-making and end-effector design. 
A subsequent study by Tinoco et al.~\cite{tinoco2021review} provides a more detailed analysis of the design of a manipulator for pruning and harvesting. The considered pruning manipulators were tested in two distinct settings: vineyards and apple orchards.
Finally, Zahid et al.~\cite{zahid2021technological} examined the technological advancements pertaining to the primary components of a robotic pruner for apple trees, including machine vision, manipulation, path planning, obstacle avoidance, and the design of the end effector.

In this work, we consider papers from the decade 2014-2024, which, as emerged from an attentive literature review, capture the evolution of trends and the recent advancements in AI that have deeply influenced this research context. For the identification of the relevant publications in this field, we carried out an extensive research using Google Scholar\footnote{\url{https://scholar.google.com/}} and Scopus\footnote{\url{https://scopus.com/}} in two main phases, the first one in June 2024 and a later one in February 2025 for recent updates and advancements.

The graph shown in Fig.~\ref{fig:histo-years} shows clearly that the number of published papers, and therefore the interest towards autonomous pruning of fruit trees and the related tasks, has consistently increased in recent years. In fact, although this field of study remains a niche area, as the number of studies indicates, it is evident that recent advancements in technology and computer vision strongly encourage the development of autonomous pruning. 
We believe that recent advancements in AI and robot autonomy have paved the way for consistent research in robotic pruning, as shown in the graph. Indeed, the figure clearly shows a substantial growth in papers involving Machine Learning (ML) technologies alongside the growth of total papers.
Hence, in this review, we discuss the latest contributions, trends, and open challenges related to the main aspects of pruning in orchards and vineyards.

\subsection{Paper organization} 
The paper is organized as follows. Section~\ref{sec:pruning_general} offers a wide overview of the autonomous pruning pipeline, giving insight into a possible subdivision of the tasks. Section~\ref{sec:visual_perception} reports visual perception methods for robotic pruning, organizing them in classical computer vision methods, in Subsection~\ref{subsec:classical-cv}, and deep-learning-based methods, in Subsection~\ref{subsec:ml-cv}. Section~\ref{sec:skeletonization} describes skeletonization and tree reconstruction methods, while Section~\ref{sec:point_estimation} describes the pruning points estimation methods. Section~\ref{sec:simulation} treats simulation environments for autonomous pruning training and development. Section~\ref{subsec:building} discusses the various hardware parts needed to constitute the different parts of an autonomous robotic pruner, starting from sensors, in Subsection~\ref{sec:sensors}, then manipulators in Subsection~\ref{subsec:manipulators}, and end-effectors in Subsection~\ref{subsec:end-eff}, and how to control them, in Subsection~\ref{subsec:planning_control}. Then, an overview of works that merge all the different parts is given in Section~\ref{sec:complete}. Finally, Section~\ref{sec:discussion} discusses the challenges and limitations, in Subsection~\ref{subsec:challenges}, and gives possible future developments in Subsection~\ref{subsec:future}. Conclusions are drawn in Section~\ref{sec:conclusions}.

\section{Autonomous Pruning with Robots}\label{sec:pruning_general}


\begin{figure*}[!ht]
    \centering
    \includegraphics[width=1\linewidth]{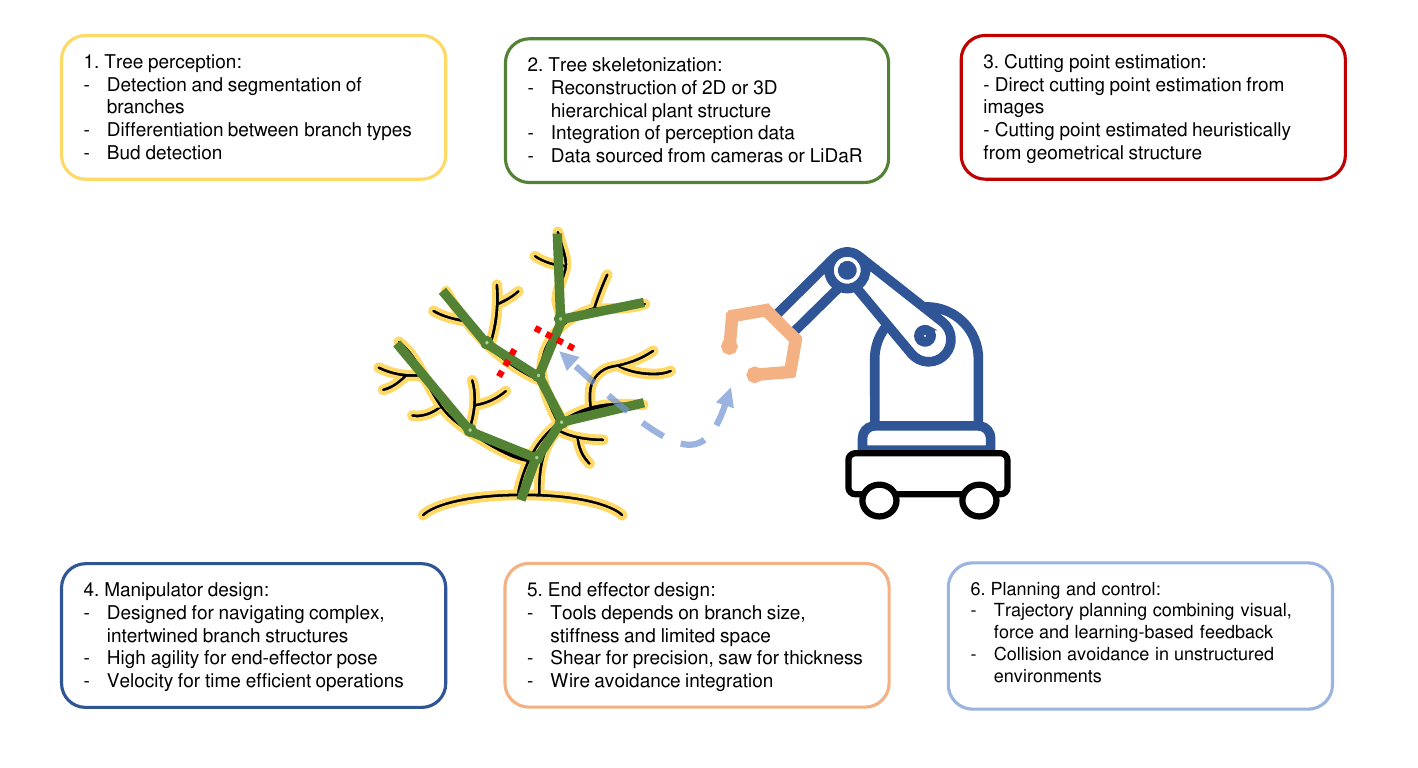}
    \caption{Scheme of the autonomous robotic pruning pipeline. The above part of the scheme depicts the perception-related tasks, while the bottom part shows the tasks related to manipulator design and control.}
    \label{fig:schema}
\end{figure*}

Automated robotics solutions are emerging as a significant support for human jobs in various sectors. Precision agriculture is one of them since it comprises many repetitive tasks in the field that require only partial human input and supervision. 
Plant pruning in orchards and, above all, vineyards represents one of the most attractive tasks in this direction. The advantages of a precise robotic solution lies on the requirement of having a fast but non-disruptive method to prune plants for new season. The intelligence of a robotic solution resides in the combination of sensors and algorithms. In fact, according to the literature of the last ten years, a robotic pruning system can be divided into a few modules, as shown in the scheme in Fig.~\ref{fig:schema}.

The robotic pruning pipeline can be decomposed in two main challenging aspects. First, a complete and efficient perception pipeline, which aims at sensing and reconstructing the structure of the target plant and its surroundings, as shown in the upper part of the scheme in Fig.~\ref{fig:schema}. Then, the design of the manipulator, its actuation system, and logic are shown in the bottom part of the figure.
As shown in the upper part of the scheme, it is possible to decompose the perception part of the pruning task into three main phases:
\begin{enumerate}
    \item \textbf{Tree perception}: sensor data, mostly obtained from cameras or LIDARs, must be interpreted to identify regions of interest. Segmentation and detection algorithms are used to distinguish the plant parts from the background. In some cases, the different plant parts, which are fundamental to have more meaningful information on the pruning action, are differentiated from each other. Some works also target the detection of the single buds lying on the branches.
    \item \textbf{Tree skeletonization}: after the regions of interest are identified, it is fundamental for the pruning operation to obtain a clear structure of the plant. For this operation, the previously obtained information about the position of the branches, their type, and some additional parameters such as their width and their orientation, is merged to form a graph-like representation, obtaining a simplified essential structure of the plant, which is more easily interpretable from the following algorithms.
    \item \textbf{Cutting point estimation}: this phase represents the crucial point of the perception activity, aiming at finally identifying the points on the plants that need to be cut. For this purpose, two main approaches are employed: the first one exploits the previously obtained structure and applies several algorithms or heuristics to find the desired cutting points. The second one consists of learning-based approaches that train algorithms based on annotations of experts.
\end{enumerate}
Therefore, the output of the perception pipeline is the pose in the 6D space of the desired cutting point, which is used as input to the manipulation part.

Regarding the manipulation, we considered three different macro-areas that were faced separately in the analyzed works:
\begin{enumerate}
    \setcounter{enumi}{3}
    \item \textbf{Manipulator design}: the architecture of the manipulator should be chosen wisely to address the movement in an intricate environment, such as interwined branches. In fact, it should have enough degrees of freedom to make the desired pose of the end-effector possible. Moreover, to be efficient enough, alongside the precision, the manipulator should be fast enough during its movement
    \item \textbf{End effector design}: the tools should be designed depending on the target branch stiffness and size, considering also the limited space due to the intricate branches and sustains. Moreover, depending on the type of cut to be made, a shear architecture has to be chosen for more precise cuts of smaller branches, while for larger ones a saw is preferrable. In some cases, due to the presence of support wires, the shears can be equipped with a specific solution for their avoidance.
    \item \textbf{Planning and control}: finally, given the input from the perception pipeline, a trajectory should be planned, considering collision avoidance. Hybrid control strategies can be adopted, combining visual force and learning-based feedback.
\end{enumerate}

\begin{figure*}[!t]
    \centering
    \includegraphics[width=1\linewidth]{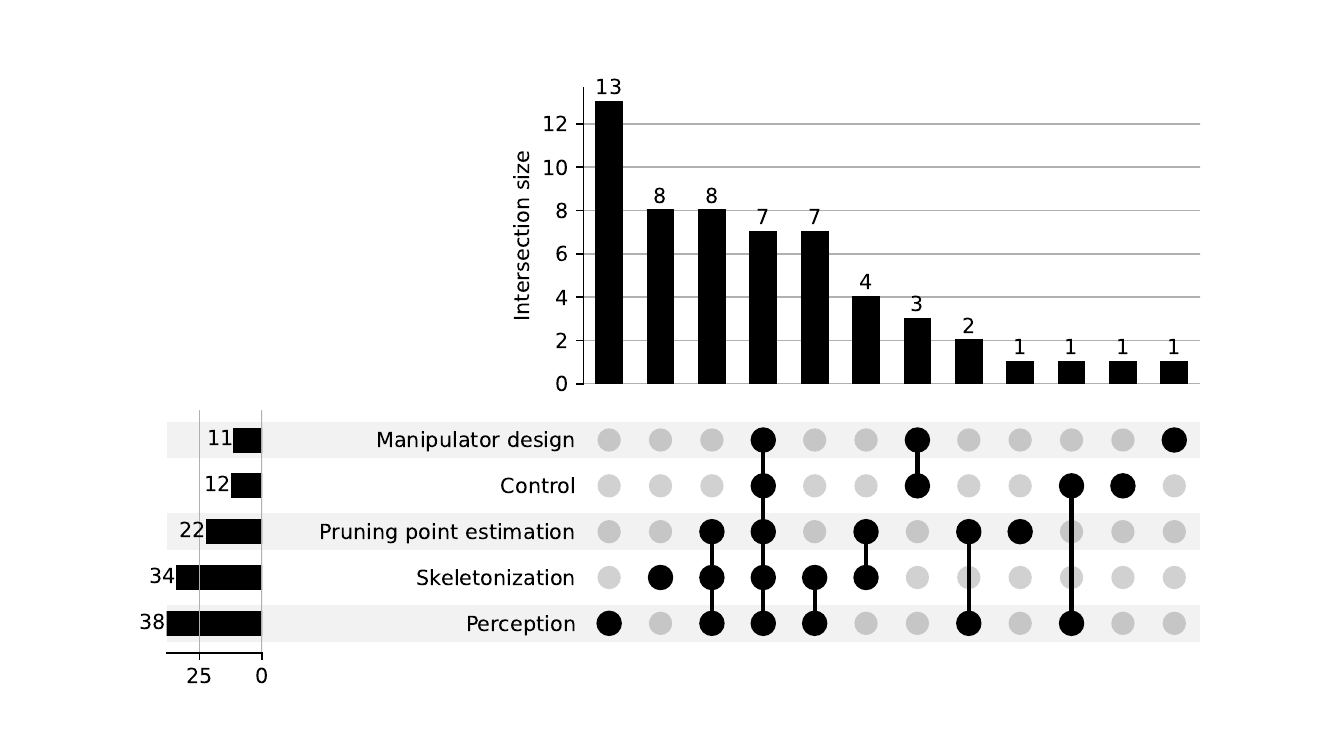}
    \caption{This plot shows the number of papers addressing individual tasks in the autonomous pruning pipeline (bottom left) and combinations of tasks (top). The manipulator and end-effector design tasks are grouped together, as they are consistently paired in the reviewed papers.}
    \label{fig:upset-plot}
\end{figure*}

The plot in Fig.~\ref{fig:upset-plot} shows the number of papers analyzed covering the different areas proposed in the previous summarization. Without losing generality, manipulator and end effector design have been merged, since they are often coupled. As the graph shows, plant perception alone is the most covered task, and it has been strongly pushed by the advancements of ML in the last five years. Right after, the skeletonization module alone and merged with the perception and the estimation of the pruning point. These trends highlight that the most covered topics are all related to the problem of sensing the plant. 
Complete research works proposing a full robotic pruning system integrating sensing and manipulation have also been published, with a similar effort in terms of the number of papers. 
Finally, the design of the manipulators, end effectors, and control has a smaller coverage in terms of published works, probably due to the more general state-of-the-art solutions already available and the recent developments of enhanced commercial platforms that can be deployed directly.

\section{Visual Perception for Robotic Pruning}
\label{sec:visual_perception}

Accurate detection and segmentation of tree structures—such as trunks, branches, and nodes—are critical tasks for enabling automation in horticultural operations like pruning. Over the years, a wide range of vision-based approaches have been developed to tackle these challenges in unstructured agricultural environments. This chapter provides an overview of the evolution of these methods, from early classical computer vision algorithms to modern deep learning-based techniques. While early efforts relied on handcrafted features and traditional image processing pipelines, recent advancements in machine learning and deep neural networks have enabled more robust and scalable solutions. The following sections outline this progression, highlighting the key contributions, technologies, and limitations of each approach.

\subsection{Classical Computer Vision Algorithms}
\label{subsec:classical-cv}
Early works strongly rely on the employment of classical computer vision algorithms, as deep learning had not yet gained traction, and the hardware lacked sufficient computational power.
Initial systems commonly used 2D RGB cameras to detect tree structures.
For instance, Qiang et al.~\cite{qiang2014identification} employed an RGB camera and morphological features with a multi-class Support Vector Machine (SVM) to detect citrus branches thicker than 5 pixels. However, they did not report recognition accuracy. Ji et al.~\cite{ji2016apple} applied a Contrast-Limited Adaptive Histogram Equalization (CLAHE) technique to segment apple tree branches, finding it superior to Otsu and basic histogram-based methods in terms of recognition rate. Shalal et al.~\cite{shalal2015orchard} combined RGB imagery with laser scanning to detect apple trunks. However, these 2D camera systems were insufficient for precise spatial localization, which limited their effectiveness in automation tasks such as pruning~\cite{gao2006image, mcfarlane1997image}.

To overcome these limitations, researchers began incorporating 3D imaging technologies, such as stereo vision and Time-of-Flight (ToF) cameras. These systems significantly improved spatial accuracy in detecting trunks, branches, and canopies. For instance, Zhang et al.~\cite{zhang2018branch} used a stereo vision camera to capture RGB, depth, and index images, achieving improved branch detection accuracy. Amatya et al.~\cite{amatya2017automated} utilized a Bayesian classifier based on morphological features (e.g., orientation, length, and thickness) to distinguish between cherry branches, leaves, fruit, and background. Building on this work, Amatya and Karkee~\cite{amatya2016integration} modeled branch geometry using linear or logarithmic functions and employed cherry cluster locations to infer occluded branches, achieving enhanced segmentation even under foliage. In a different approach, Botterill et al.~\cite{botterill_robot_2017} reconstructed vine structures in 3D using a triangulation-based feature matching algorithm, followed by a Radial Basis Function SVM (RBF-SVM) classifier trained on hand-labeled $5\times 5$ image patches. To increase efficiency, 95.5\% of the pixels were processed using a simpler SVM model, with only the remaining pixels classified by the more complex RBF-SVM. Controlled lighting and monochrome backgrounds further improved segmentation performance in their setup.

\subsection{Deep Learning Techniques 2D}
\label{subsec:ml-cv}

\begin{figure*}[!t]
\centering
\subfloat[]{\includegraphics[width=0.4\textwidth]{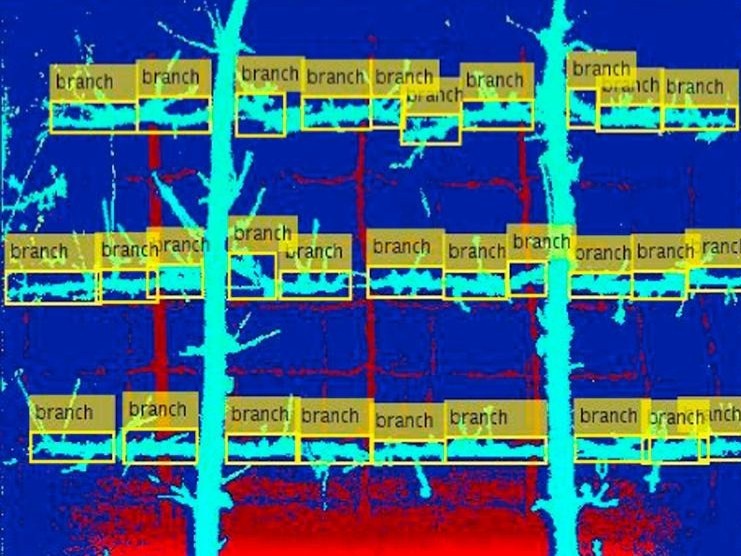}%
\label{fig:det-zhang2018}}
\hfil
\subfloat[]{\includegraphics[width=0.4\textwidth]{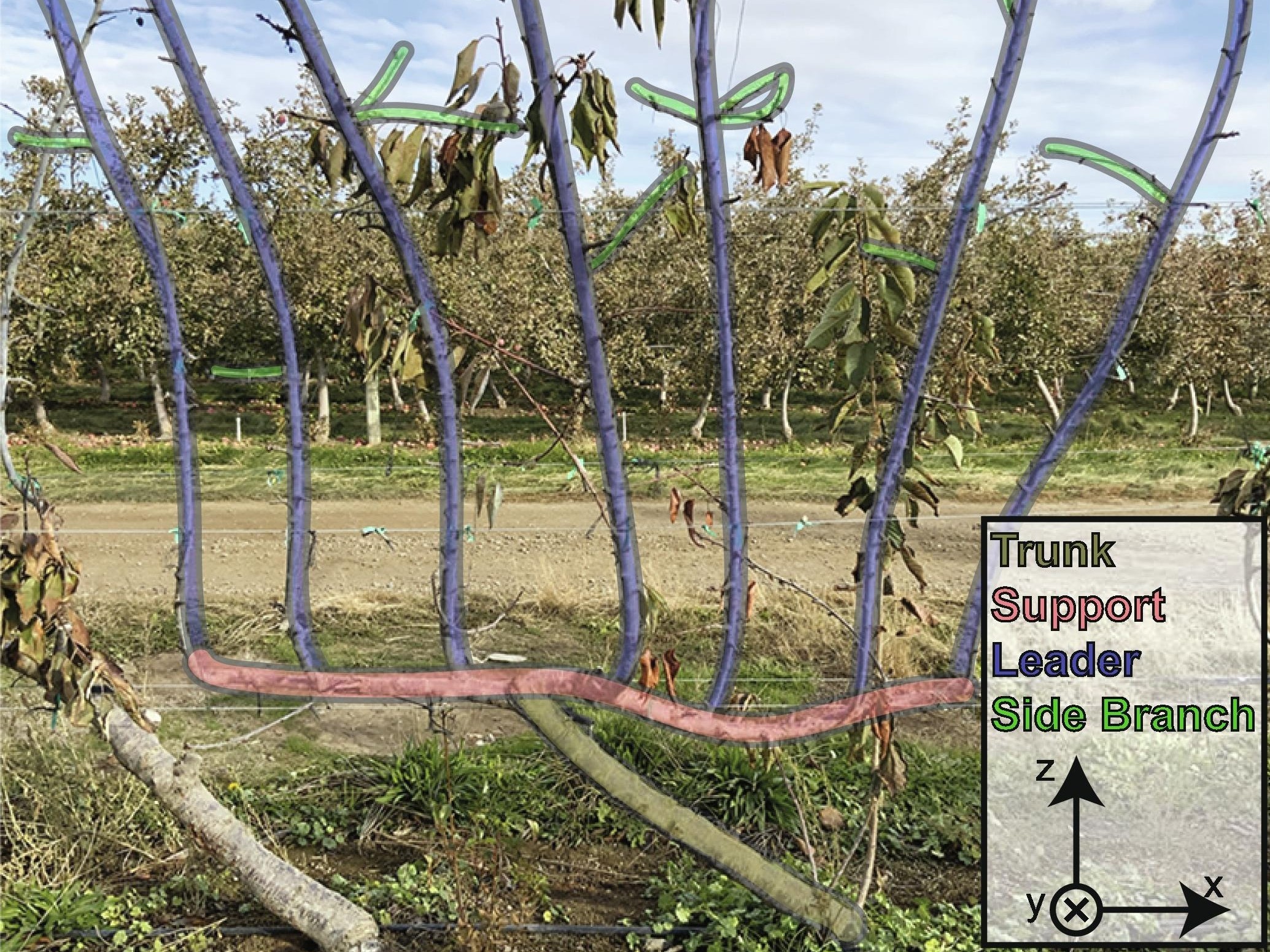}%
\label{fig:det-y0u2022}}
\hfil
\subfloat[]{\includegraphics[width=0.4\textwidth]{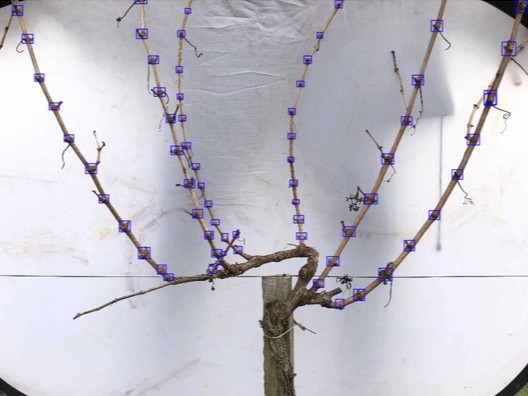}%
\label{fig:oliveira2024}}
\hfil
\subfloat[]{\includegraphics[width=0.4\textwidth]{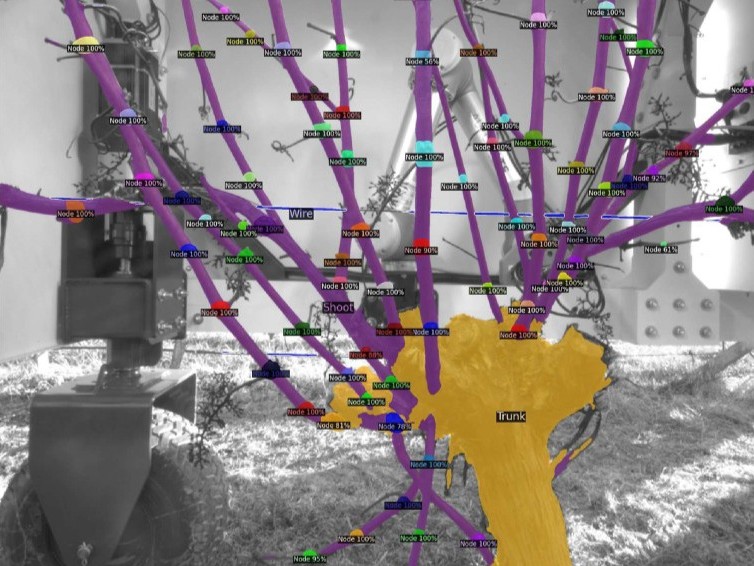}%
\label{fig:det-williams2023}}
\caption{Overview of four different tasks of computer vision for autonomous pruning: (a): detection of branches on pseudo-color image extracted from PCD~\cite{zhang2018branch}, (b): semantic segmentation of different parts of cherry trees~\cite{you2022semantics}, (c): bud detection on grapevine~\cite{oliveira2024enhancing}, (d): panoptic segmentation of the structure of different parts of grapevine and different buds\cite{williams2023modelling}.}
\label{fig:detection}
\end{figure*}

\begin{table}[!htbp]
\centering
\caption{Overview of deep learning-based computer vision methods for various tree crops.}
\resizebox{0.95\textwidth}{!}{%
\begin{tabularx}{\textwidth}{lXXcc}
\toprule
Crops & Method & Performance & Year & Ref. \\ \midrule
Citrus & SVM + morphological ops & Branch seg. (diameter >5 px) & 2014 & \cite{qiang2014identification} \\ \midrule[0.1pt]
Generic tree & HSV color + edge detection & Seg. acc. = 96.64\% & 2015 & \cite{shalal2015orchard} \\ \midrule[0.1pt]
Apple & CLAHE-based seg. & Acc. = 94\% & 2016 & \cite{ji2016apple} \\ \midrule[0.1pt]
Cherry & Bayesian classifier & Acc. = 93.8\% & 2016 & \cite{amatya2016integration} \\ \midrule[0.1pt]
Cherry & Bayesian classifier & Acc. = 89.2\% (w/ foliage) & 2017 & \cite{amatya2017automated} \\ \midrule[0.1pt]
Grapevine & RBF-SVM & P = 58\%, R = 49\% (canes) & 2017 & \cite{botterill_robot_2017} \\ \midrule[0.1pt]
Apple & R-CNN (AlexNet) & PCI: P = 86\%, R = 81\%; +Depth: P = 92\%, R = 86\% & 2018 & \cite{zhang2018branch} \\ \midrule[0.1pt]
Apple & Depth cut + SegNet (multi-class) & Trunk: Acc. = 0.92, IoU = 0.59; Branch: Acc. = 0.93, IoU = 0.44 & 2018 & \cite{majeed_apple_2018} \\ \midrule[0.1pt]
Grapevine & Faster R-CNN (various backbones) & Best (ResNet18): F1 = 0.55, AP = 45.1\% & 2019 & \cite{majeed2019study} \\ \midrule[0.1pt]
Apple & Faster R-CNN + U-net (UAV) & Det: IoU = 0.735, F1 = 0.925; Seg: Acc. = 94.68\% & 2020 & \cite{wu_extracting_2020} \\ \midrule[0.1pt]
Apple & Depth cut + SegNet (multi-class) & Trunk: Acc. = 0.91; Branch: Acc. = 0.92; Wire: Acc. = 0.97 & 2020 & \cite{majeed2020deep} \\ \midrule[0.1pt]
Grapevine & CNN (FCN-VGG16 best) & F1: background = 0.97, trunk = 0.94 & 2020 & \cite{majeed2020determining} \\ \midrule[0.1pt]
Grapevine & Mask R-CNN (ResNet101) & Avg P = 41.1\%, R = 48.7\% & 2021 & \cite{fernandes_grapevine_2021} \\ \midrule[0.1pt]
Guava & Tiny Mask R-CNN & F1 = 0.518 & 2021 & \cite{lin_three-dimensional_2021} \\ \midrule[0.1pt]
Jujubee & SPGNet + DBSCAN & Trunk: IoU = 0.85; Branch: IoU = 0.76 & 2021 & \cite{ma_automatic_2021} \\ \midrule[0.1pt]
Apple & Cascade Mask R-CNN (Swin-T) & @IoU=0.5: bboxAP = 9.879, seg. AP = 0.893 & 2022 & \cite{tong_branch_2022} \\ \midrule[0.1pt]
Cherry & FlowNet2 + pix2pix (sim) & IoU = 62.3\%, FN = 28.7\%, FP = 1.8\% & 2022 & \cite{you2022optical} \\ \midrule[0.1pt]
Cherry & FlowNet2 + pix2pix (sim) & - & 2022 & \cite{you2022precision} \\ \midrule[0.1pt]
Cherry & Mask R-CNN + postproc. & IoU: 0.78 (light), 0.67 (nat.); P = 0.89/0.81; R = 0.97/0.81 & 2023 & \cite{borrenpohl2023automated} \\ \midrule[0.1pt]
Grapevine & Faster R-CNN + Mask R-CNN & Node seg.: Acc. = 0.88, R = 0.85; pruning pt. det. = 0.97 & 2023 & \cite{guadagna2023using} \\ \midrule[0.1pt]
Apple/Grapevine & HOB-CNN (occlusion) & RMSE = 1.63–2.66, Corr. = 0.925–0.948 & 2023 & \cite{chen2023hob} \\ \midrule[0.1pt]
Apple & UNet++ (InceptionV3) & IoU = 0.625, F1 = 0.7692 & 2023 & \cite{kok2023obscured} \\ \midrule[0.1pt]
Apple & SOLOv2 (ResNet50) & AP: front = 0.934, back = 0.917, side = 0.940, cloudy = 0.947 & 2023 & \cite{tong2023image} \\ \midrule[0.1pt]
Grapevine & Hourglass net & Nodes: F1 = 0.92, ADE = 0.66px; Courson: F1 = 0.75 & 2023 & \cite{gentilhomme2023towards} \\ \midrule[0.1pt]
Grapevine & Detectron2 (panoptic) & PQ: node = 0.6832, wire = 0.674; SQ varies 0.77–0.85 & 2023 & \cite{williams2023modelling} \\ \midrule[0.1pt]
Grapevine & YOLOv7–10 & F1 = 70–86.5\% & 2024 & \cite{oliveira2024enhancing} \\ \midrule[0.1pt]
Grapevine & PSP-Net + YOLOv5 & PSP: mIoU = 83.7\%, Acc. = 96.8\%; YOLOv5: F1 = 0.72 & 2024 & \cite{chen_grapevine_2024} \\ 
\bottomrule
\end{tabularx}
}
\label{tab:tree_segmentation}
\end{table}

Adopting deep learning methods has significantly improved the accuracy of tree structure detection and segmentation. Despite the complexity of outdoor environments—characterized by occlusions, variable lighting, and unstructured backgrounds—deep neural networks have demonstrated strong generalization capabilities, enabling reliable identification of trunks, branches, and other tree components.
In the field of branch detection and segmentation, several machine learning algorithms and deep learning architectures have been proposed: a complete overview of the different employed methods is shown in Table~\ref{tab:tree_segmentation}, and some example of the possible applications of deep computer vision are shown in Fig.~\ref{fig:detection}.
In particular, Convolutional Neural Networks (CNN) have demonstrated strong performance by learning visual features such as edges, textures, and shapes that are critical for detecting trunks and branches. Their translation invariance and scalability make them well-suited to handling variations in branch positions and lighting conditions.

Region-based CNN (R-CNN)~\cite{girshick2015region} architectures have been widely applied for tree component detection and segmentation. These models rely on region proposals to localize potential objects before classifying them and generating bounding boxes or segmentation masks. Fast R-CNN~\cite{girshick2015fast} and Faster R-CNN~\cite{ren2015faster} enhance computational efficiency by processing the entire image with a CNN to generate feature maps and using a Region Proposal Network (RPN). Mask R-CNN~\cite{mask_rcnn} extends this approach by adding a segmentation mask prediction branch.

Numerous studies have adopted these architectures for different detection tasks. Tong et al.~\cite{tong_branch_2022}, for instance, evaluated two kinds of R-CNN, namely Mask R-CNN  and Cascade Mask R-CNN~\cite{cai2019cascade}, one of its variants, to detect and segment the trunk, primary branches, and supports of apple trees. For each model, they tested two backbones: Swin-T~\cite{liu2021swin} and ResNet50, showing that when Intersection over Union (IoU) is set to 0.5, Cascade Mask R-CNN with Swin-T backbone achieves the best result.
Borrenpohl and Karkee~\cite{borrenpohl2023automated} compared two models of a Mask R-CNN with a ResNet101 backbone to detect vigorous leaders from Upright Fruiting Upshots (UFO) in sweet cherry trees, both in artificial and natural light. Images were first segmented using a stereo image and a deep stereo matching~\cite{cheng2020hierarchical}, to identify the tree of interest, and then manually annotated to identify the trunk and branches. The accuracy was increased by developing two post-processing algorithms: the first to avoid double detections of the same leader by comparing the IoU of two different masks in the images, while the second to avoid simultaneous detection of two adjacent leaders, showing how active light provides higher scores. 
In other cases, such R-CNN are used to segment multiple grapevine branch classes, such as main cordon, canes, and nodes, which facilitates the process of reconstructing the tree structure and the following estimation of pruning points. In this regard, Fernandes et al.~\cite{fernandes_grapevine_2021} compared a Mask R-CNN with several backbone models (R50-FPN, R101-FPN, X101-FPN), demonstrating ResNet101 to be the best backbone. 
Similarly, Guadagna et al.~\cite{guadagna2023using} fine-tuned a Mask R-CNN for grapevine organ segmentation, distinguishing more classes: cordon, arm, spur, cane, and node. The trained model was then tested on images of light-pruned, shoot-thinned, and unthinned control samples. In this case, better performances were achieved on thinned shoots compared to unthinned control samples, underlining the role of canopy management in improving the performance of autonomous solutions.
A tiny Mask R-CNN was also used by Lin et al.~\cite{lin_three-dimensional_2021} to simultaneously obtain an instance segmentation of branches and fruits in guava crops for the sake of obstacle avoidance during fruit harvesting. In fact, after detecting fruits and branches, they were reconstructed as spheres and cylindrical segments starting from the segmented frames and the point cloud data.

Other works proposed the usage of a detection model, such as the one by Majeed et al.~\cite{majeed2019study}, where a Faster R-CNN was employed to detect the visible part of the cordon for green shoot thinning in grapevine. It was deployed through transfer learning and fine-tuning using pre-trained networks such as AlexNet, VGG16, VGG19, and ResNet18, obtaining a higher accuracy with the latter.
A Faster R-CNN has also been used by Wu et al. ~\cite{wu_extracting_2020} to identify apple trees' bounding boxes from UAV images. Then, a U-net model~\cite{robbeberger_unet_2015}, which was conceived for medical applications, was applied on the previously extracted bounding boxes to segment branch pixels. It allowed the extraction of a convex-hull polygon~\cite{graham1972efficient} for each tree, to estimate several vegetative indexes such as the Asymmetry Index, Roundness Index, and Compactness Index.

In similar applications, R-CNN has been employed with different inputs: Zhang et al.~\cite{zhang2018branch}, which addressed the problem of branch detection in apple orchards using a Pseudo-Color depth Image (PCI) obtained by mapping the depth to a certain color value, as shown in Fig.~\ref{fig:det-zhang2018}. The adopted architecture consisted of an improved AlexNet~\cite{krizhevsky2012imagenet} network trained and compared on depth images and PCIs and depth images together. 
The issue about the reconstruction of the position of occluded has been raised by Chen et al.~\cite{chen2023hob}, where a regression-based deep learning model, Hallucination of Occluded Branches CNN (HOB-CNN), is used for this purpose in apples and vine plants, obtaining the three structures directly. Tree branch position prediction is formulated as a regression problem towards the horizontal locations of the branch along the vertical direction or vice versa. Two different versions of U-Net~\cite{robbeberger_unet_2015} were trained for the reconstruction of the trunk: U-Net Visible is used to detect only visible branch sections, while U-Net Whole is used to hallucinate the whole branch in occlusion conditions. Finally, a curve fitting method is employed to obtain a single-pixel skeleton of the plants.
Similarly, Kok et al.~\cite{kok2023obscured}, used a U-net++~\cite{zhou2019unet++} with a InceptionV3 encoder to segment visible apple branches, followed by a skeletonization algorithm in order to reconstruct the obstructed parts of the plant.

SegNet~\cite{badrinarayanan2017segnet}, an encoder-decoder fully convolutional network, has been used extensively for semantic segmentation. Majeed et al.~\cite{majeed_apple_2018} applied SegNet on RGB and point cloud data to segment apple trunks and branches, achieving strong precision and boundary-F1 scores. In a follow-up study~\cite{majeed2020deep}, they extended the segmentation to trellis wires, confirming improved performance when foreground-RGB images were used. Majeed et al.~\cite{majeed2020determining} also compared SegNet, AlexNet, and VGG-based fully convolutional networks for segmenting bilateral cordons in grapevines, combining segmentation with a centroid-based mathematical model to distinguish right and left cordons.

Another popular architecture employed in different work is SegNet~\cite{badrinarayanan2017segnet}, a fully convolutional network for semantic pixel-wise segmentation consisting of an encoder-decoder structure.
Majeed et al. (2018)~\cite{majeed_apple_2018} employed RGB images and Pointcloud Data (PCD) to detect one-year-old apple trees. After a preprocessing employing the depth information, to cut out the background noise, SegNet was employed for the semantic segmentation of background, trunk and branches. 
In a follow-up study~\cite{majeed2020deep}, they extended the segmentation to trellis wires, confirming improved performance when foreground-RGB images were used. 
In another work, Majeed et al.~\cite{majeed2020determining} segmented the two cordon on grapevines, comparing SegNet with its weight initialized with VGG16 and VGG19~\cite{simonyan2014very}, and other fully convolutional networks, such as AlexNet and VGG16 modified for semantic segmentation as in~\cite{long2015fully}. Moreover, to separate the right and the left cordons, a mathematical model was applied, evaluating first the centroid of the trunk and, later, the two parts of the cordon and fitting a curve on the centroids of each cordon to better fit their shape.

Other segmentation models were successfully employed. Tong et al.~\cite{tong2023image} used a SOLOv2~\cite{wang2020solov2} with a Resnet50 backbone to segment RGB images of dormant apple trees, identifying trunks, branches, and supports. It is then compared to A Mask R-CNN and Cascade Mask R-CNN, achieving a better performance, also compared in different illumination weather conditions. 
Moreover, Chen et al.~\cite{chen_grapevine_2024} employed a PSP-net~\cite{zhao2017pyramid} to separate branches and the main stem from the background in grapevines. The results show better performance with respect to other considered models such as U-net and DeepLabv3+. Moreover, on the segmentation mask achieved with the first step of segmentation, different versions of YOLO were employed to detect the nodes of the canes. Experimental results show how YOLOv5 achieves better results. 
Similarly, Oliveira et al.~\cite{oliveira2024enhancing} compared four different versions of YOLO, such as YOLOv7, YOLOv8, YOLOv9, and YOLOv10, for the detection of nodes in dormant grapevine. They trained the networks using a public dataset presented in~\cite{gentilhomme2023towards} with images presenting diverse artificial backgrounds, and validated them on other data, contributing with a new dataset. YOLOv7 was assessed as the most robust model, achieving a trade-off between accuracy and inference speed. The bud detection task can be seen in Fig.~\ref{fig:oliveira2024}.
With the same objective, Gentilhomme et al.~\cite{gentilhomme2023towards} employed an architecture based on a Stacked Hourglass Network~\cite{newell2016stacked}, ViNet, which outputs a heatmap with the different kinds of branches of interest such as nodes and branches. It produces a stack of node and branch heatmaps. The position of the nodes is extracted through a post-processing step that localizes local maxima within large enough blobs in the heat maps.
Williams et al.~\cite{williams2023modelling}\cite{williams2024archie} developed a novel vision system, which aims at reconstructing the whole structure of the plant. The detection system employs a Detectron2~\cite{wu2019detectron2} for panoptic segmentation~\cite{kirillov2019panoptic} of the cane information from the 2D images, as seen in Fig.~\ref{fig:det-williams2023}. In this case, panoptic segmentation is used for semantic segmentation of trunk, canes, and wires, while for the instance segmentation of the different nodes, which are fundamental to reconstruct the structure of the plant.

You et al.~\cite{you2022optical} exploit the optical flow technique joined with neural networks to enhance the segmentation of branches, trained entirely in a simulation environment as already shown in a previous work~\cite{you2022precision}. First, optical flow is estimated with a fully-sized FlowNet2 with pre-trained weights~\cite{ilg2017flownet}. Then, it is colorized and, alongside the three-channel RGB frame, is passed to a pix2pix~\cite{isola2017image} Generative Adversarial Networks to perform segmentation of the branches. Moreover, a neural network comparison is carried out with different types of inputs, demonstrating the similar accuracy demonstrated by the simulation environment and the real world, demonstrating significantly more consistent and robust performance.

Most of the aforementioned models operate on 2D RGB data. A limited number of models directly exploit 3D data such as depth images or point clouds, which, in the case of robotic pruning, may offer some additional information that is fundamental for a better 3D reconstruction of the considered plant. 
For instance, Ma et al.~\cite{ma_automatic_2021} utilized two cameras to obtain a high-quality 3D point cloud in the field, involving a ToF sensor. A reconstruction pipeline was developed to create a comprehensive 3D model from just two perspectives and, finally, they used the deep learning method SPGNet~\cite{landrieu2018large} to automatically segment trunks and branches.
However, in most cases, the depth information is added a posteriori, once the branches have been detected during the structure reconstruction phase, as seen in the following chapter.

\section{Skeletonization and Tree Structure Reconstruction}
\label{sec:skeletonization}

\begin{table*}[!htbp]
\centering
\scriptsize
\setlength{\tabcolsep}{4pt}
\caption{Overview of skeletonization and thinning methods.}
\resizebox{\textwidth}{!}{%
\begin{tabular}{lp{4.5cm}p{4cm}cc}
    \toprule
Crop & Method & Performances & Year & References \\
\midrule
Generic tree & SbG + RANSAC & Branch Id. Acc. = 96.0\%, Branch modeling Acc. = 92.2\%, Radius Acc. = 78.0\% & 2015 & \cite{elfiky2015automation} \\ \midrule[0.1pt]
Apple tree & Semicircle fitting + Center point optimization & Diameter Acc. = 89.4\% (±5mm), Branch ID = 100\% & 2016 & \cite{chattopadhyay2016measuring} \\ \midrule[0.1pt]
Grapevine & Stochastic Grammar + RDP line detection & Overlap with groundtruth: 95\% & 2017 & \cite{botterill_robot_2017} \\ \midrule[0.1pt]
Apple tree & Split-and-merge + robust cylinder fitting & Branch ID Acc. = 96\%, Diameter error = 0.6 cm & 2017 & \cite{medeiros_modeling_2017} \\ \midrule[0.1pt]
Apple tree & Polynomial interpolation & Corr. Coeff. r = 0.91 & 2018 & \cite{zhang2018branch} \\ \midrule[0.1pt]
Apple tree & Harris corner detection + AHCA & - & 2019 & \cite{liu_research_2019} \\ \midrule[0.1pt]
Grapevine & 6-th degree polynomial interpolation & 80\% R\textsuperscript{2} > 0.98 & 2020 & \cite{majeed2020determining} \\ \midrule[0.1pt]
Grapevine & Auxiliary matrix & - & 2021 & \cite{fernandes_grapevine_2021} \\ \midrule[0.1pt]
Guava tree & Point cloud + RANSAC cylinders & 2D F1 = 0.394, 3D F1 = 0.451 & 2021 & \cite{lin_three-dimensional_2021} \\ \midrule[0.1pt]
Apple tree & 2D junction point estimation & F1 = 91.06\% & 2022 & \cite{tong_branch_2022} \\ \midrule[0.1pt]
Cherry & DBSCAN + Space Colonization & Angle RMSE = 3.309°, Trunk RMSE = 0.069 m & 2022 & \cite{xu2022improved} \\ \midrule[0.1pt]
Cherry & Topo/Geom. Constraints & Median Acc. = 70\% & 2022 & \cite{you2022semantics} \\ \midrule[0.1pt]
Apple tree & Point2Skeleton + OBR & Total err. = 18.68 ± 14.3 mm, Obscured = 38.11 ± 32.64 mm & 2023 & \cite{kok2023obscured} \\ \midrule[0.1pt]
Apple tree & 3D junction estimation & Diam. MAE = 1.10 mm, Spacing MAE = 16.06 mm & 2023 & \cite{tong2023image} \\ \midrule[0.1pt]
Grapevine & Shortest-path on resistivity graph & - & 2023 & \cite{gentilhomme2023towards} \\ \midrule[0.1pt]
Generic tree & DBSCAN + slicing & Pos. dev. = 0.418 cm, Dir. dev. = 8.47° & 2023 & \cite{you2023tree} \\ \midrule[0.1pt]
Jujubee & ICP + SCA & Reg. err. = 0.66–0.91 cm & 2023 & \cite{fu2023skeleton} \\ \midrule[0.1pt]
Grapevine & Zhang–Suen + YOLOv5 & Thinning = 95.44\%, Sens. = 90.87\% & 2024 & \cite{chen_grapevine_2024} \\ \midrule[0.1pt]
Walnut & Delaunay + Dijkstra + MST + LM Fit & Acc. dev. < 7\% & 2024 & \cite{li2024efficient} \\ \midrule[0.1pt]
Grapevine & 2D/3D node constraints & Conn. Acc.: 67.05\% (1-side), 78.4\% (2-side) & 2023 & \cite{williams2023modelling} \\ 
\bottomrule
\end{tabular}
}
\label{tab:skeletonization}
\end{table*}

After the tree is detected, the data obtained from the detection phase must be interpreted and exploited to reconstruct its structure. The reconstruction of the tree's shape can be performed in 2D or 3D, merging depth data from stereo cameras or other sources. A broad overview about all the analyzed methods is shown in Table~\ref{tab:skeletonization} For this scope, Duki\'{c} et al.~\cite{dukic2024branch}, presented the BRANCH dataset, a multiple-viewpoint dataset of 70 images which covers pear trees, both before and after pruning. They proposed a pointcloud extraction for the multiple POV images, then the 3D final model is obtained by using the RANSAC and Iterative Closest Point (ICP)~\cite{besl1992method} algorithms. In addition, by superimposing the pointcloud before and after the pruning, they obtained the position of the pruned branches.
A similar multi-POV approach has been adopted by Li et al.~\cite{li2024efficient}, where the PCD of a ToF sensor have been merged and filtered to reconstruct the full structure of a walnut tree. Moreover, the IWOA-RANSAC-NDT algorithms are used for the 3D registration. Then, Delaunay triangulation~\cite{su2022adaptive} and Dijkstra shortest path~\cite{straub2022approach} algorithms are employed to evaluate the minimum spanning tree, a concept originating from the nutrient transport paths in ecology. Then, the Kd-tree data structure is used for branch segmentation and, finally, the Levenberg-Marquardt algorithm~\cite{hu2020non} is used to obtain the full skeleton model.

An important part of the tree reconstruction process consists of the estimation of the parameters of the tree, such as in the work by Tabb and Medeiros~\cite{tabb2017robotic}, where parameters such as branch diameters, branch length, and branch angles are estimated. 
Another more recent study, by Dong et al.~\cite{dong2024improved}, introduces a novel method for three-dimensional trunk and branch volume calculation of individual apple trees, enabling the creation of pre- and post-pruning maps using a voxel-based algorithm.

\begin{figure*}[!t]
    \centering
    \includegraphics[width=0.9\textwidth]{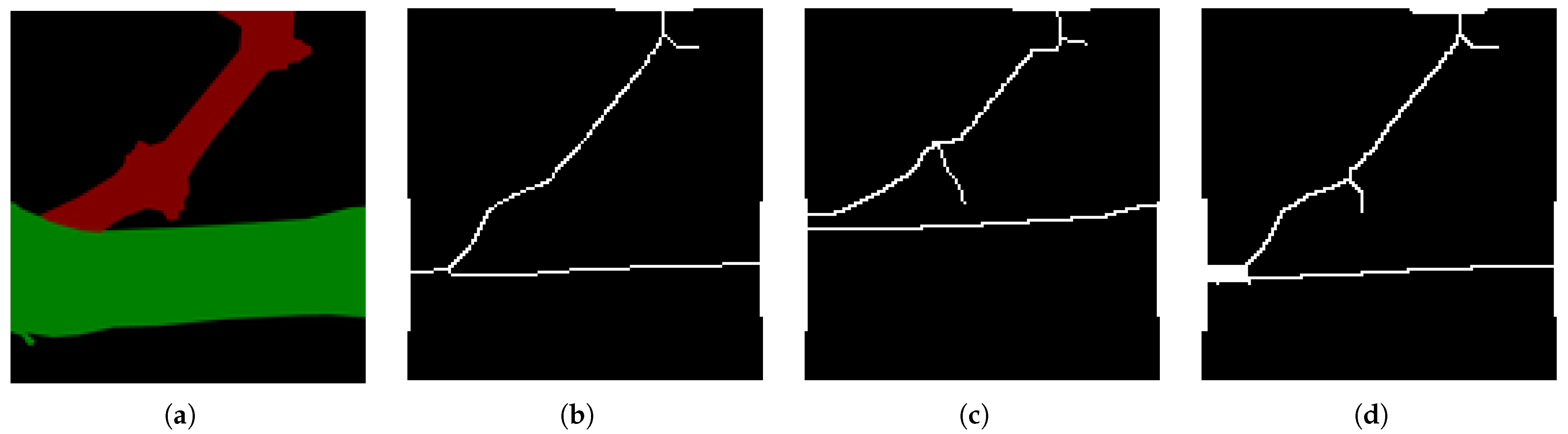}
    \caption{Comparison with Zhang and Suen with other methods~\cite{chen_grapevine_2024}. (a) RGB image. (b) The result image of the Zhang–Suen thinning algorithm. (c) The result image of the Hilditch algorithm. (d) The result image of the Stentiford algorithm.}
    \label{fig:thinning-zhang-suen}
\end{figure*}

Some studies use simple solutions, such as Zhang et al.~\cite{zhang2018branch}, where a polynomial fitting is used as a skeletonization algorithm after the segmentation mask is obtained. 
Majeed et al.~\cite{majeed2020determining} reconstructed the structure of the cordon in a vine plant by fitting its centroids, obtained through applying an 8-connectivity matrix to the segmentation mask, with several methods such as polynomial, Gaussian, Fourier and sum of sines. Also, Borrenphol et al.~\cite{borrenpohl2023automated} employed a high-degree polynomial to fit each prediction of branches of cherry tree to estimate each trajectory, differentiating one from another.
Another study by Botterill et al.~\cite{botterill_robot_2017} starts from the a-priori knowledge that each cane is a smooth curve with approximately uniform thickness and they are all connected to a tree. For this sake, they employed a Stochastic Image Grammar~\cite{zhu2007stochastic} to represent this knowledge, which is described in~\cite{botterill2013finding}. First, the cane edge segments are extracted by taking the outline of the foreground and approximating them as a set of straight lines using the Ramer-Douglas-Peucker~\cite{ramer1972iterative} algorithm. The nodes are joined through an SVM considering lengths, thickness, curvature, and distances.

The approach proposed by Zhang and Suen~\cite{zhang_fast_1984} found several applications for skeletonizing tree structures. It consists of a fast parallel thinning algorithm consisting of two subiterations:  the first targets the removal of southeast boundary points and northwest corner points, while the second focuses on eliminating northwest boundary points and southeast corner points. This approach keeps endpoint and pixel connectivity unchanged, thinning each pattern to a skeleton of unitary thickness.
In fact, a comparison with other skeletonization algorithms for the sake of plant structure reconstruction has been carried out by Cuevas et al. for rose structure reconstruction~\cite{cuevas_segmentation_2020}, demonstrating its superiority.
For instance, Tong et al.~\cite{tong_branch_2022} were able to obtain the skeleton structure of the apple trees once they evaluated the 2D segmentation mask, reconstructing the junction points between the trunk and the primary branches. In this case, the junction point between the trunk and the branches has been evaluated at the intersection of the main trunk and branches in the skeleton, taking into account the diameter of the branches. The same algorithm has also been employed in~\cite{tong2023image}, identifying the three-dimensional structure of the tree with respect to the camera. If the difference between the z-coordinates of the two end-points of the branch skeleton line is greater than a certain threshold, the line between the two endpoints is considered a junction point. Moreover, in this case, the pruning point estimation is based solely on the geometrical estimation of the diameter of the branches, at a certain distance from the trunk.
Also Chen et al.~\cite{chen_grapevine_2024} uses the same algorithm for plant skeletonization reconstruction, after it emerged to be the best one when compared with the Hilitch thinning algorithm~\cite{hilitch1969linear} and the Stentifod thinning algorithm~\cite{martin2000image}, joined with buds identification with a YOLOv5 network: the final obtained data is added the depth information, which results in the 3D reconstruction of grapevine plants. The comparison can be seen in Fig.~\ref{fig:thinning-zhang-suen}.

To reconstruct the skeleton line, Ma et al.~\cite{ma_automatic_2021} employed a Laplacian algorithm~\cite{cao2010point} and, after downsampling, calculated their eigenvectors by a Fat Point Feature Histogram (FPFH). Then, a Sampling Consistency Algorithm (SAC-IA) was used to achieve the mapping relationship between the skeleton points. Finally, the classical ICP algorithm was applied to refine the initial matrix. After the application of a network for branch-type segmentation. A clustering algorithm, Density-Based Spatial Clustering of Applications with Noise (DBSCAN), was used to estimate the branch count.
You et al.~\cite{you2023tree} employed Terrestrial Laser Scanning (TLS) to reconstruct the tree skeleton. First, DBSCAN is employed to remove all the outliers of the obtained point, recording the point-traversal order of each point. Then, the point cloud was divided in slices employing contour planes, obtaining several tree segments by applying DBSCAN. After the point-inversion transformation, the tree skeleton points were obtained from each tree segment. The relationship between skeleton points and the weight of each part was evaluated based on the point-traversal order, organizing the branch hierarchy. Angle consistency was used to optimize the skeleton structure, employing positional and directivity deviations.
Similarly, Xu et al.~\cite{xu2022improved} employed terrestrial laser scanning o and Space Colonization Algorithm (SCA) on cherries and begonia trees to obtain their skeletal structure. Since the SCA algorithm alone leads to over-extension of the skeletal trunk of the trees, improvements are made, preprocessing the data with noise filtering and identifying the number of bifurcations using the DBSCAN algorithm.

Liu et al.~\cite{liu_research_2019} reconstructed a graph representing the spatial structure of apple branches, and employed a Harris' corner detection algorithm~\cite{harris1988combined} to process binary images to detect the initial candidate point. Then, a 2-level Aggregation Hierarchy Clustering (2HAC) is used to address invalid points in the initial candidates, and finally skeleton points are obtained. An E-line converging algorithm is proposed for processing to obtain an adjacency matrix.
Also, Fu et al.~\cite{fu2023skeleton} employed a 3D pointcloud, reconstructed from frames taken from different points and merged with the ICP algorithm, as in~\cite{fu2020three}. Once the full point cloud has been reconstructed, the space colonization algorithm~\cite{runions2005modeling} has been employed to reconstruct the structure of the plant, where neighboring space points of the skeleton influence the structure to be formed. The results of this process consist of a directed graph, whose nodes are later classified heuristically to separate the trunk from primary and secondary branches. The obtained model consists of multi-segment cylindrical-connection graphs, considering the decreasing diameters of branches.
You et al.~\cite{you2022semantics} aim to reconstruct the skeleton of a UFO cherry tree from a point cloud in the form of a graph defined by edges, nodes, and labels, depending on the type of branches. In this process, the point cloud is first pre-processed, downsampling it into superpoints. A set of graph edges is identified using nearest neighbors, and each edge's likelihood to belong to the skeleton is evaluated. Finally, branch tips are identified. Then, to identify the best skeleton candidate, an unlikely standard skeletonization algorithm, some a-priori knowledge about the structure of the threes in employed to formulate some constraints used to score the best candidates. In fact, some assumptions are made based on the label-based topological constraints, which rely on the label progression, label linearity, trunk-support split, and label-based geometric constraints, which are based on turning angles and growth direction.
Also, Fernandes et. al.~\cite{fernandes_grapevine_2021} represent the structure of a grapevine as a graph. Starting with the multiclass segmentation of cordon, canes, and nodes, they identify three sets of connections in the graph structure: the one connecting cordon to canes, the interconnection between elements of the same canes, and the connections between nodes and canes. The algorithm employed is based an auxiliary matrix containing all main cordon instances. The algorithm identifies the main cordon overlapping a cane mask by checking non-zero values and unique overlaps. If no overlap is found, the cane mask is dilated incrementally until an overlap is detected or a limit is reached. Parameters like dilation size, number of dilations, and vertical slots are user-defined.

Several applications have been exploiting several geometry fitting algorithms, approximating trunk and branches to geometric figures such as cylinders.
Elfiky et al.~\cite{elfiky2015automation} investigated the use of Skeleton-based Geometric (SbG) to reconstruct the 3D skeleton of plants. A point cloud is obtained by merging different perspectives. First, a Laplacian smoothing is performed to extract the skeleton of the tree. Once the graph is obtained, an algorithm based on Geometric Feature Extraction is presented to assign the nodes respectively to the trunk or to the branches. Then, circles are recursively fitted to the point cloud, estimating the best locations for the centers of the circles and the radii, depending also on adjacent circles, using a Random Sample Consensus-based (RANSAC) algorithm and fitting the circles on the 2D plane when the pointcloud is split on horizontal planes. Primary branches are then modeled as cylinders using RANSAC, individuating their growing direction and segmenting them to differentiate them from the other branches.

Akbar et al.~\cite{akbar2016novel} and Chattopadhyay et al.~\cite{chattopadhyay2016measuring} built a 3D reconstruction of apple trees exploiting the predominantly cylindrical structure of the branches, introducing a new public dataset for benchmarking. In fact, the diameters and the centers of the branches are estimated from a single depth image, filtering lens distortion, and background. Then, the skeleton of the plant is reconstructed~\cite{cao2010point}, the neighboring points of a skeleton segment are found, the dominant direction is found, the neighboring points are projected into the cross section, and a semicircle is fit on it, optimizing the center points~\cite{marquardt1963algorithm}. The results showed that the proposed scheme provides a performance of 89\% for correctly estimating the diameter of the primary branches.
Meideiros et al.~\cite{medeiros_modeling_2017} employed a laser sensor from a different perspective to measure and model the fruit tree. At first, points are classified in three sets (trunk, junction point, and branches) by a split-and-merge clustering. Then, trunk candidates and junction points are refined by a robust fitting algorithm that models trunks and branches as cylinders, measuring the diameters of the trunk and of the primary branches.
In another case, Lin et al.~\cite{lin_three-dimensional_2021} reconstructed the shape of branches and fruits in guava plants under the hypothesis where guava fruit detections could be approximated to a sphere and irregular branches to a finite sequence of 3D cylindrical segments. Within this hypothesis, a RANSAC sphere fitting method~\cite{schnabel2007efficient} and a Principal Component Analysis (PCA) were investigated to reconstruct fruit and branches from the point cloud, once the region of interest is identified via segmentation. 
Kok et al.~\cite{kok2023obscured} obtained the a representation made by a series of skeletal points and the radius by employing a PointNet++~\cite{qi2017pointnet++} with a shared multilayer perception network~\cite{lin2021point2skeleton}, an unsupervised method that learns the skeletal representation by utilizing insights from the Medial Axis Transform (MAT) to capture the intrinsic geometric and topological characteristics of the point cloud. To connect skeletal points a greedy algorithm inspired by~\cite{yandun2020visual}, connecting the points hierarchically in the bottom-up direction. Moreover, this work also introduces an algorithm to recover branches obscured by vegetation, namely Obscured Branch Recovery (ORB).

\begin{figure}[!ht]
    \centering
    \includegraphics[width=0.9\linewidth]{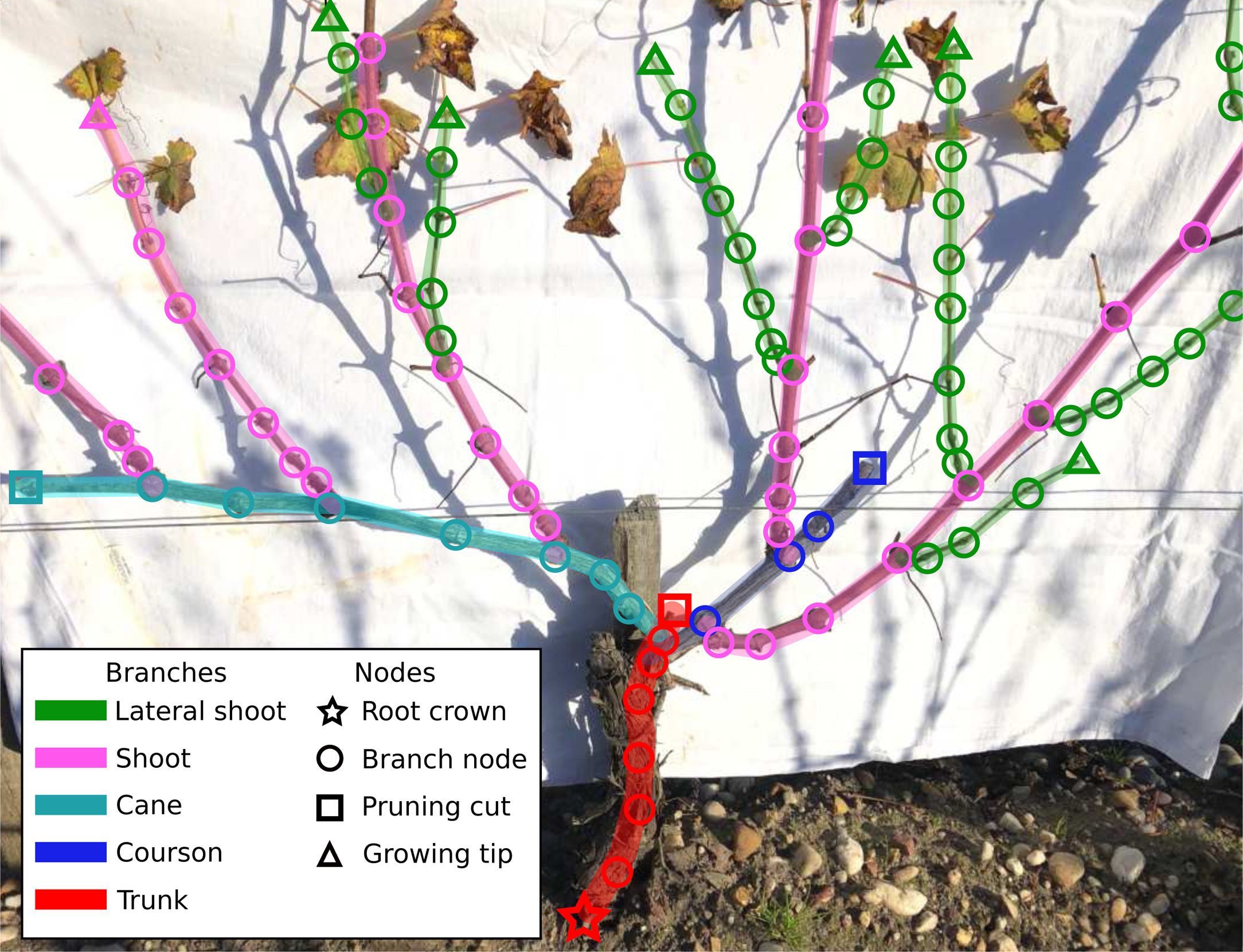}
    \caption{Example of target grapevine plant structure. The types of branches are shown, such as branches and nodes, including different branch types (depicted in different colors), as well as the structure. The dots represent the plant nodes~\cite{gentilhomme2023towards}.}
    \label{fig:example-structure}
\end{figure}

Gentilhomme et al.~\cite{gentilhomme2023towards} start from a set of nodes of different types and belonging to different branch categories, and connect them to build the grapevine structure. They rely on the predicted branch vector fields to derive the spatial dependencies between nodes. It is highlighted how finding an optimal association by analyzing all the possible connections is an NP-hard problem. Studies about human pose estimation~\cite{cao2010point} optimize it by creating a spanning tree skeleton and matching adjacent body parts, assuming known node patterns. Here, with unknown nodes and patterns, the approach instead uses shortest-path optimization on a resistivity graph to link nodes to parent nodes, forming a tree structure that’s robust and adaptable to varying patterns and missing data. An example of the desired output is shown in Fig.~\ref{fig:example-structure}.

Williams et al.~\cite{williams2023modelling} exploit the geometric feature of the grapevine, asserting that the internode cane growth occurs similarly to a straight line. Taking advantage of this, several images from different perspectives were taken, using a Charuco board as a background for calibration, and the 3D structure was recreated by imposing the following constraint. First, the lines between two nodes fitting 95\% of their length in cane segmentation were kept. Then, depth data is used to determine if the lines are contiguous in 3D space. Discontinuities are assessed when neighbouring pixels on an image are more than 2 mm apart. Finally, canes were linked to complete the final structure of the vine, eliminating invalid connections.

\section{Pruning point estimation}
\label{sec:point_estimation}

\begin{table*}[!ht]
    \caption{Overview of pruning point estimation method}
    \centering
\resizebox{\textwidth}{!}{
\begin{tabularx}{\textwidth}{lXXcc}
\toprule
Crop       & Method                                                                                          & Performances                                                                 & Year                         & Ref.   \\ \midrule
Grapevine  & Cost function depending on parameters (length, position, angle from head, etc)                & -                                                                           & 2017     & \cite{botterill_robot_2017}                     \\ \midrule[0.1pt]
Grapevine  & Midpoint between nodes. The final point is located above the second node of a cane.                & -                                                                           & 2021 & \cite{fernandes_grapevine_2021}                 \\ \midrule[0.1pt]
Grapevine  & Faster R-CNN for detecting pruning regions,                                                     & Recall =  (76±15) \%, Precision =  (82±12)\%, F1-score = (78±11)\%             & 2021      & \cite{guadagna202116}                        \\ \midrule[0.1pt]
Apple tree & Estimation of branches diameter                                                                 & Accuracy = 87.2\%                                                            & 2023             &  \cite{tong2023image}                  \\ \midrule[0.1pt]
Jujubee    & Skeleton analysis: 1/3 of the lenght of the primary branch                                      & -                                                                           & 2023           & \cite{fu2023skeleton}                   \\ \midrule[0.1pt]
Grapevine  & Faster R-CNN for detecting pruning regions,                                                     & Highest pruning point detection (depending on visibility conditions) = 0.97 & 2023       & \cite{guadagna2023using}                    \\ \midrule[0.1pt]
Grapevine  & Faster R-CNN for detecting pruning regions (depending on wood type, orientation and visibility) & Highest detection rate: visible intermediate complex spurs = 0.97           & 2023        & \cite{guadagna2023using}\\ \midrule[0.1pt]
Grapevine  & Analysis of the skeleton and the buds position                                                  & Accuracy = 82.35\%                                                           & 2024      & \cite{chen_grapevine_2024}      \\
\bottomrule              
\end{tabularx}
}

    \label{tab:pruning-points}
\end{table*}

\begin{figure}[!t]
    \centering
    \includegraphics[width=0.9\linewidth]{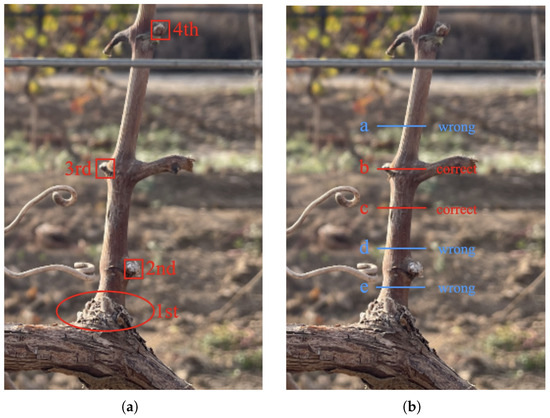}
    \caption{Example of how the pruning point is evaluated depending on the position of the buds on a grapevine.~\cite{chen_grapevine_2024}}
    \label{fig:pru-example}
\end{figure}

\begin{figure*}[!t]
\centering
\subfloat[]{\includegraphics[width=0.4\textwidth]{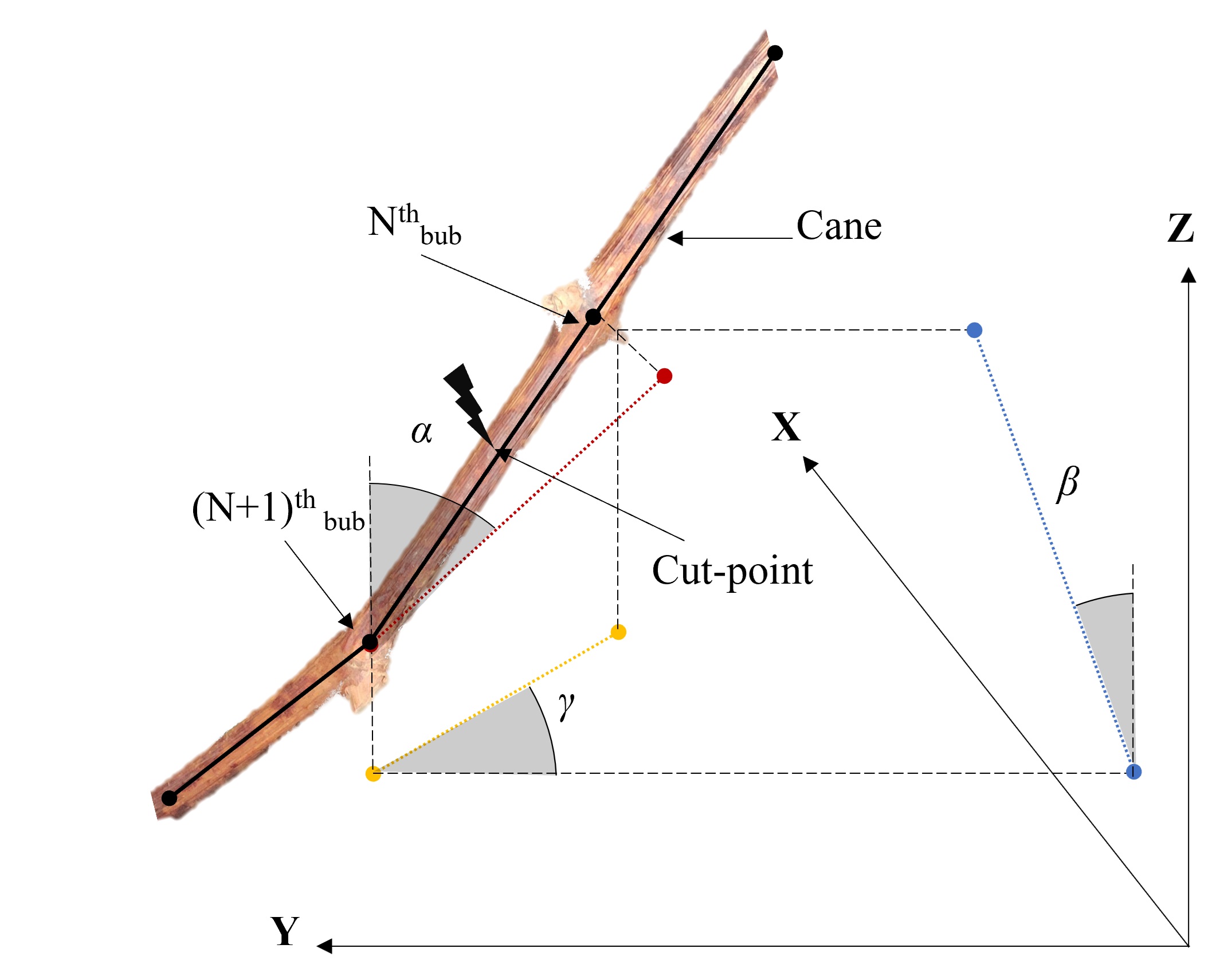}%
\label{fig:prun-silwal2022}}
\hfil
\subfloat[]{\includegraphics[width=0.29\textwidth]{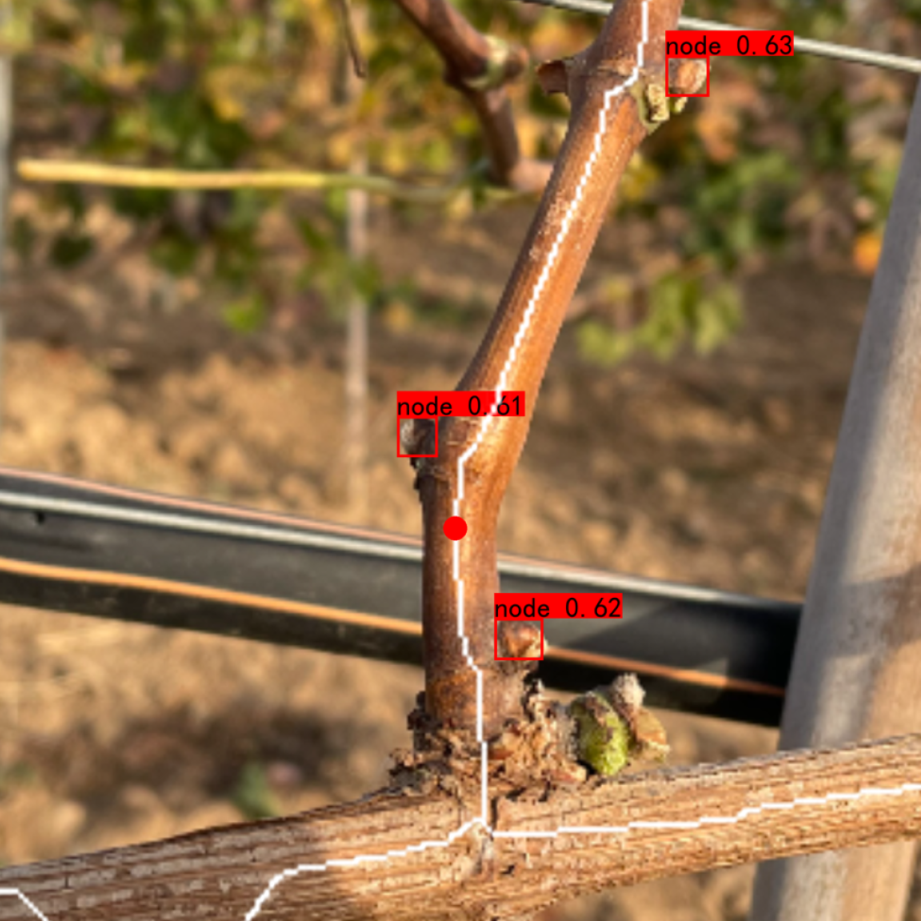}%
\label{fig:prun-chen2024}}
\hfil
\subfloat[]{\includegraphics[width=0.29\textwidth]{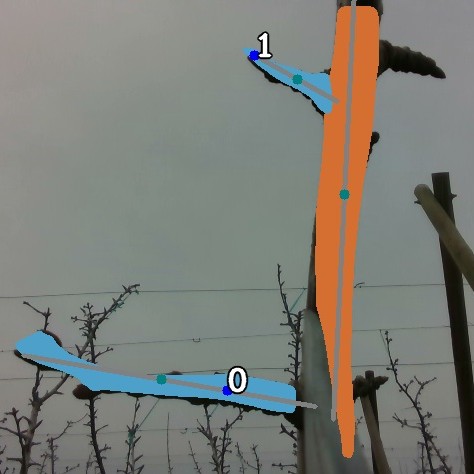}%
\label{fig:prun-you2022}}

\caption{(a): cut-point detection: three-dimensional vector projection of a cane section in the YZ (red
line – roll angle $\alpha$), ZX (blue line – pitch angle $\beta$) and XY axis (yellow line – yaw angle $\gamma$)~\cite{silwal2022bumblebee}. (b): correct identification of pruning points on grapevine~\cite{chen_grapevine_2024}. (c): correct identification of pruning points on cherry tree~\cite{you_autonomous_2022}.} 
\label{fig:pruning-point-estimation}
\end{figure*}

In the previous sections, several aspects of the process of identifying and reconstructing a tree's structure have been discussed. In fact, the robustness and reliability of the tree detection and structure estimation process are fundamental to accurately estimate pruning points. 
Analyzing the works presented in this section, two primary methodologies emerge: the first one relies on the previously obtained structure. In contrast, the second one is based on training algorithms directly on examples provided by experts. Fig.~\ref{fig:pruning-point-estimation} provides an example of correct estimation of pruning points in different trees.

Generally, the localization of pruning points can be seen as a combinatorial optimization problem, as claimed by the work of Strnad et Kohek~\cite{strnad2017novel}. Here, two novel discrete differential evolution (DDE) methods are proposed to optimize simulated tree pruning. In this case, the objective function to be optimized can be defined by an empirical model of light interception. The two proposed methods rely on a path-based and a set-based DDE, which are closely related to a discrete search domain under the implemented mutation operators. Within this problem configuration, the main issue is defining a quantifiable set of metrics to be evaluated and optimized. For this purpose, Williams et al.~\cite{williams2023modelling} proposed a new set of quantifiable metrics, named DOLPHIN, which aim at evaluating the quality of a grapevine cane, to provide a criterion for pruning. The set of metrics includes the diameter of the cane, measured between the second and third nodes, the orientation of the cane with respect to the center of the vine's head, length, position on the head, health of the cane, internode length, and node count along the cane. These parameters can help to measure the quality of a cane to be kept or cut for the next year, and could also be applied to other kinds of crops.
In other cases, the policy for the decision of the branches to be pruned can be more shallow: for instance, Fu et al.~\cite{fu2023skeleton} considered that in the jujube plant, the primary branches should be shortened each year by approximately 1/3, to reduce nutrient consumption. Therefore, given the previously obtained plant graph, the evaluation of the pruning point is based only on the branch length. Other works consider a more complex vector of features (such as length, position, angle from head, distance below wires, growth position with respect to the trunk) to decide which grapevines' canes to keep. A cost function, represented by a simple linear combination of features, was parametrized using vines labeled by experts measuring pruning quality~\cite{kolmanivc2021algorithm}.

In other plants, such as the grapevine, it is necessary to consider a more complex structure, differentiating the various types of branches. In this case, a previous semantic segmentation and branch type identification, with a successive reconstruction, is fundamental. Moreover, bud detection also plays a significant role~\cite{marset2021towards}\cite{diaz2018grapevine},  since the position of the cutting point depends strongly on their position, as shown in Fig.~\ref{fig:pru-example}. Fernandes et al.~\cite{fernandes_grapevine_2021}, after reconstructing the structure of the plant, obtain the potential pruning points between two nodes on the same cane or between the bases of two canes growing from the same cane, between the base of a cane and its first node.
Similarly, Chen et al.~\cite{chen_grapevine_2024}, after reconstructing the structure of the grapevine, identify the position of the buds, merged with the depth data, to better identify the pruning point. In fact, the step followed to achieve this result consists of a connection check, where bud pixel coordinates are linked via the mask map to group buds on the same branch, and a key point selection, where the bud with the highest vertical coordinate is chosen as the key point. Starting from the key point, the pruning point is identified on the skeleton image, and its 3D coordinates are recorded.

In other works, the pruning points are directly estimated without previously reconstructing the structure of the plant, training the estimation algorithms directly on experts' annotations.
For instance, Guadagna et al.~\cite{guadagna202116} trained a Faster R-CNN~\cite{ren2015faster} model for identifying pruning regions in vineyards on a hand-labeled dataset. Then, the obtained result is analyzed according to the type of wood, orientation, and visibility of the pruning point. 
In a later work, Guadagna et al.~\cite{guadagna2023using} fine-tuned a Faster R-CNN with images collected in various vineyards and categorized according by wood type, orientation and visibility. It was emphasized that the accuracy of detecting pruning regions was strongly influenced by visibility.

\section{Simulation Environments}
\label{sec:simulation}

In the field of autonomous pruning system research, the use of simulation is of great importance for several reasons. First, it provides a safe environment for evaluating the algorithms' effectiveness, given their destructive nature and the potential risk of compromising the health of the plants. In addition, simulation allows training data generation for the proposed algorithms, which is often time-consuming and difficult to obtain with real-world data. Nevertheless, the generation of a realistic three-dimensional structure of branches is a significant challenge in the field of three-dimensional reconstruction~\cite{zhao_multiple_2024}\cite{tang2015integrated}. In fact, in order to obtain a more accurate model, the processes influencing the evolution of plants should be taken into account~\cite{kolmanivc2021algorithm}, such as the growth against gravity (gravitropism)~\cite{berut2018gravisensors}, the search for light (phototropism)~\cite{mvech1996visual} and the competition with other plants~\cite{palubicki2009self}.

One of the earliest simulated environments was created by Corbett-Davies et al.~\cite{corbett-davies_expert_2012}, where a system for making pruning decisions using simulated vines is presented. It employs a vector space model where decisions are made based on a vector of characteristics of a cane. An expert evaluated the approach by comparing it to the pruning decisions of a typical human pruner: it performed similarly to humans in $89\%$ of the cases and better in $30\%$ of the cases.
The realism of plant models is essential in simulation environments: the accuracy directly affects the system's ability to scale to real-world applications, directly improving the transferability of skills or algorithms to real-world environments.
In addition, simulations must provide a variety of scenarios to increase the robustness of training. This diversity helps expose models to various conditions, improving their adaptability and reducing the risk of overfitting, ultimately leading to more reliable real-world performance.
To this end, Zhao and Wang~\cite{zhao_multiple_2024} used a Space Colonization Algorithm to generate 3D skeletons of cherry trees~\cite{runions2007modeling}. The generated models were then segmented with different labels (trunk and branches), and the annotated point cloud was exported for training neural networks for tree structure detection and pruning, and point estimation. Some constraints inherent to SCA were used to obtain more realistic plant structures, including potential growth directions. 
In addition, Bryson et al.~\cite{bryson2023using} illustrate that artificially generated point clouds from a custom tree simulator can achieve comparable results when used to train detection algorithms. In particular, a PointNet++ network trained on the synthetic dataset outperformed a network trained on a real dataset acquired with a LiDAR, with an increase in IoU of approximately $1-7\%$.

\begin{figure*}[!t]
\centering
\subfloat[]{\includegraphics[width=0.49\textwidth]{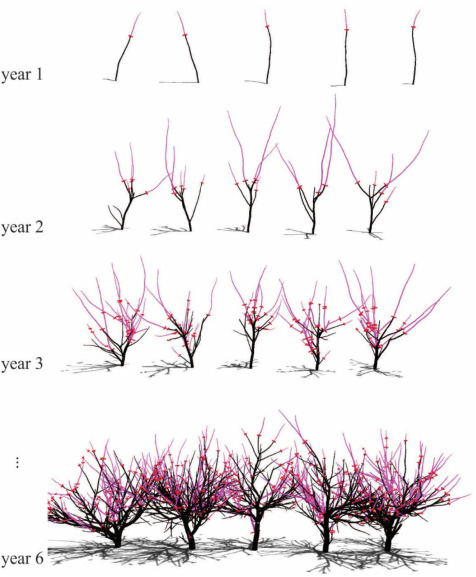}%
\label{fig:sim-kolmanvic2021}}
\hfil
\subfloat[]{\includegraphics[width=0.49\textwidth]{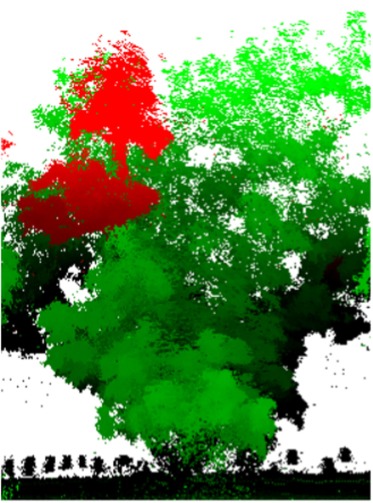}%
\label{fig:sim-westling2021}}
\caption{Two possible applications of simulation environments. (a): evolution of the growth of the plant after pruning during several years in apple trees\cite{kolmanivc2021algorithm}. (b): estimation of the tree crown eliminated given a pruning point on a mango tree~\cite{westling_procedure_2021}.}
\label{fig:sim}
\end{figure*}

Kolmanivc et al.~\cite{kolmanivc2021algorithm} demonstrated the possibility of a simulation environment by developing an algorithm for automated apple tree pruning in simulation, based on creating artificial tree structures. The authors used the previously developed framework EduAPPLE~\cite{kohek2015eduapple}, which is based on hierarchical modularity, as many other simulators~\cite{kang2016imapple, de1988plant}. In addition, phenomena such as phototropism and competitive growth are included in the simulation, in addition to the evaluation of the amount of light received by buds, as demonstrated in~\cite{palubicki2009self}, making it possible to evaluate the impact of pruning in subsequent years, as shown in Fig.~\ref{fig:sim-kolmanvic2021}. In this work, a novel algorithm based on differential evolution~\cite{storn1997differential} optimizes the light received by the inner parts of the plant and minimizes the shading to achieve the optimal result. Subsequently, two pruning templates, namely cylinder and cone, were used to compare the proposed method and expert pruning, resulting in a comparable performance. 

Simulation is also a valuable tool for visualizing immediately the effects of pruning on the structure of the plant, as highlighted by Westling et al.~\cite{westling_procedure_2021}, who proposed a LiDaR-based method for suggesting pruning strategies on avocado and mango plants. This method is based on light availability on different plant parts and uses ray tracing to create a voxelized tree structure. Next, the tree was reconstructed as a graph and then explored using the A* algorithm~\cite{hart1968formal}. Using a structural analysis of the tree, the simulation environment determined which portion of the point cloud, corresponding to a plant part, is pruned, based on a single, specific cut point, as shown in Fig.~\ref{fig:sim-westling2021}. Finally, novel cut points were proposed by identifying tree components that negatively affect the light distribution.

In addition, simulation environments can help improve the understanding of plant structure, as in the work of Qiu et al.~\cite{qiu20243d}, where the incompleteness and discontinuity of point clouds were improved using a closed-loop framework~\cite{qiu20223d_characterization}. In fact, the issue of PCD incompleteness is highlighted as a limitation, and it was observed that modern orchards with high tree density result in high occlusion and low sensor visibility. In addition, adverse weather conditions, such as variable sunlight and strong winds, introduce noise and ghost points, further degrading the quality of the data. To overcome these challenges, it is necessary to reconstruct complete point clouds starting from raw data accurately. This will facilitate the use of branch data from individual trees for a comprehensive robotic pruning strategy.
To tackle this problem, a closed-loop framework is proposed,  consisting of integrating a data generation process based on real-world observations (Real2Sim) and a subsequent application of these data in a simulation environment (Sim2Real). Three-dimensional apple tree models were generated through a characterization pipeline~\cite{qiu20223d_characterization}, which was used to train a skeletonization model. Subsequently, a transformer encoder~\cite{yu2023adapointr} and a joint decoder using a Generative Adversarial Network provide point feature embedding and coarse completion of the point cloud. In addition, the performance of Sim2Real was evaluated by comparing branch-level feature characterization errors using incomplete raw data and complete data.

\section{Building a robotic platform for autonomous pruning}
\label{subsec:building}
The advancement of autonomous pruning systems for orchards and vineyards requires carefully selecting and designing the manipulator and end effector. These elements are critical to ensuring efficient and precise operation. The manipulator must be tailored to the specific pruning task. It must possess the necessary degrees of freedom, strength, and dexterity to navigate complex and variable environments, such as intertwined branches and irregular plant structures~\cite{tinoco2021review}. Furthermore, it must be capable of sophisticated obstacle avoidance, allowing it to maneuver through dense foliage and tight spaces without causing damage to the surrounding vegetation~\cite{yang2017obstacle}. It is equally crucial to ensure the end effector is designed with the highest level of accuracy to perform precise, clean cuts that promote healthy regrowth. These two components must work in harmony to enable reliable, effective, and safe pruning in autonomous agricultural systems. There is a clear gap in the literature regarding the employment of a mechanical actuator and the development of a control system that can plan an efficient path, avoid obstacles, and move the end effector toward the goal. Indeed, numerous similar works concentrate on the autonomous harvesting of fruits in orchards and vineyards~\cite{bac2017performance}\cite{harrell1990robotic}\cite{silwal2017design}\cite{tanigaki2008cherry}\cite{zhang2015kinematics}. In this section we will analyze in detail the study of manipulators, end-effectors and, path planning and control regarding the development of an autonomous pruner.

\subsection{Sensors for Robotic Pruning}
\label{sec:sensors}

\begin{figure}[!t]
    \centering
    \includegraphics[width=0.75\linewidth]{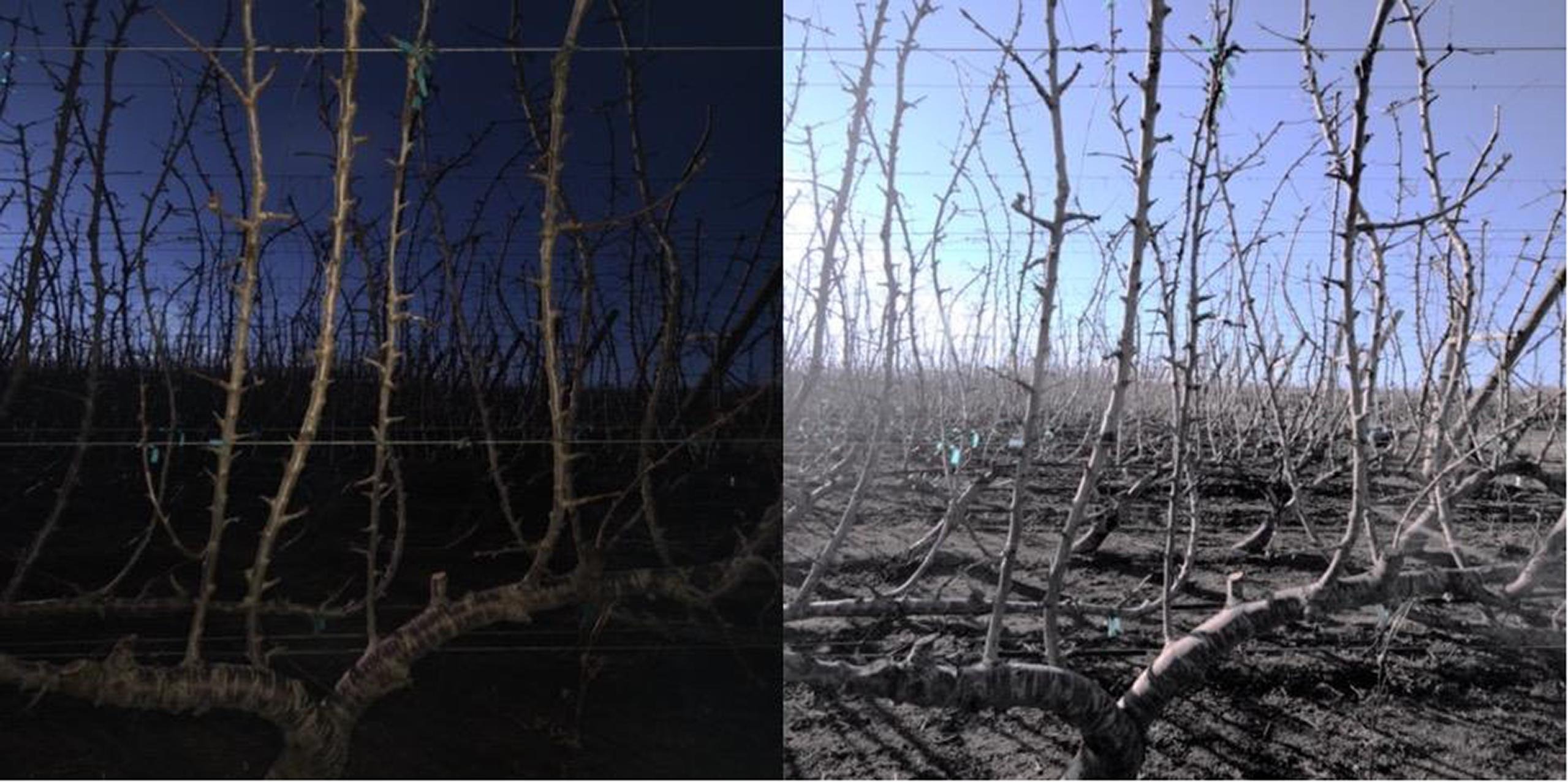}
    \caption{Comparison of natural and artificial light~\cite{borrenpohl2023automated}.}
    \label{fig:artificial-light}
\end{figure}

Autonomous pruning of orchards and vineyards relies on a variety of advanced sensors to ensure precision and efficiency. 
Different types of cameras have been used to gather datasets of RGB frames for branch detection and segmentation. A consistent group of works utilizes a reflex and high-resolution RGB camera to capture images of branches, followed by a second phase of segmentation. This sensor's main advantage is its high resolution, which is ideal for identifying branches, even the smaller secondary ones. However, high-resolution images will inevitably result in a lower frame rate in some algorithms due to the large quantity of data to process. Some studies highlight how the illumination influences the performance of the considered algorithms, comparing the orientation of the sun and the weather conditions. Moreover, in some cases, as shown in Fig.~\ref{fig:artificial-light}, artificial light is employed, demonstrating to be a valid approach to obtain a uniform and controlled illumination.

Many works utilize RGB-D cameras, such as the Intel RealSense or Kinect family. These cameras provide an estimation of the depth of the frame, despite the slightly lower adjustable resolution. They are stereo cameras, which is why they are the best choice for this task. This conversion to a point cloud provides meaningful information for segmentation, reconstruction of the branches, identification of the pruning points, and control system. However, it has some limitations. The limited measured interval and sparse depth resolution may pose a significant problem at certain thresholds due to the invisibility of the smallest branches and buds.

Obtaining accurate depth information for small objects, such as branches, presents a considerable challenge for stereo vision systems. In fact, depth images derive from a disparity map, used to infer depth information from two slightly different images (stereo images) captured by cameras placed side by side. The limited number of pixels representing these objects makes it difficult to perform the necessary comparisons between stereo images for reliable depth calculation. Additionally, the common occurrence of occlusions further exacerbates this problem by preventing the establishment of correct correspondences. The resulting inaccuracies in depth estimation can have significant consequences in applications relying on precise spatial understanding of the environment, particularly in scenarios where the fine details of small objects are crucial.

In some cases, traditional stereo vision systems are replaced or complemented with structured light or time-of-flight (ToF) sensors, which are the optimal choice for providing more reliable depth information for small and challenging objects. In other works, such as~\cite{ma_automatic_2021}, multiple cameras are used to obtain a more accurate point cloud reconstruction. Specifically, they positioned two Azure Kinect DK cameras at a fixed distance and synchronized them with an audio cable and artificial illumination. They used an ArUco calibration board at approximately 1.5 m in front of the camera to register the point cloud. Another work by Li et al.~\cite{li2023automatic} analyzed the planting mode to better define the way of image acquisition. They identified the challenges of matching images of dormant jujube trees, specifically the lack of local feature regions, weak continuity, and variations in individual morphological structures. They developed a bilateral image-matching method based on three-view geometry constraints to address this. EXIF tags enhanced reconstruction efficiency by implementing a camera self-calibration algorithm based on images. This enabled them to create a 3D point cloud using a Structure from Motion (SfM) reconstruction strategy, combined with a Patch-based Multi-view (PMVS) algorithm.

\subsection{Manipulators}
\label{subsec:manipulators}
The classification of manipulators depends on the number of joints, which corresponds to the number of degrees of freedom (DoF), and the joint type, which is either revolute (R) or prismatic (P). These two variables have a significant impact on the manipulator's key properties, including dexterity, obstacle avoidance, spatial requirements, and the final orientation of the end-effector, which is essential for this task.
The literature provides only a limited number of examples of manipulators used for robotic pruning. Table~\ref{tab:manipulator} shows a summary of the different manipulators taken into consideration in this paper.

\begin{figure*}[!t]
\centering
\subfloat[]{\includegraphics[width=0.39\textwidth]{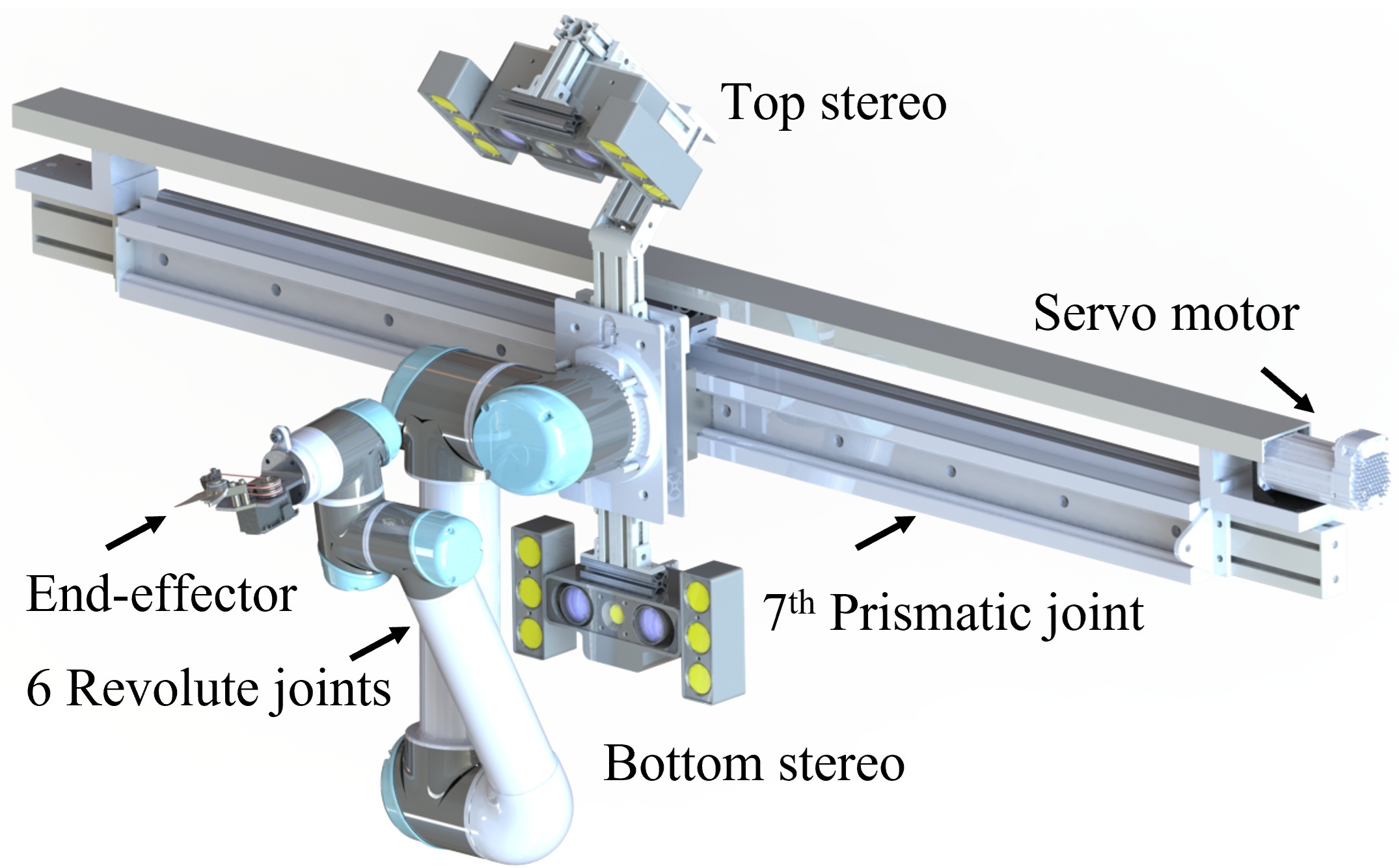}%
\label{fig:bumblebee-manipulator}}
\hfil
\subfloat[]{\includegraphics[width=0.59\textwidth]{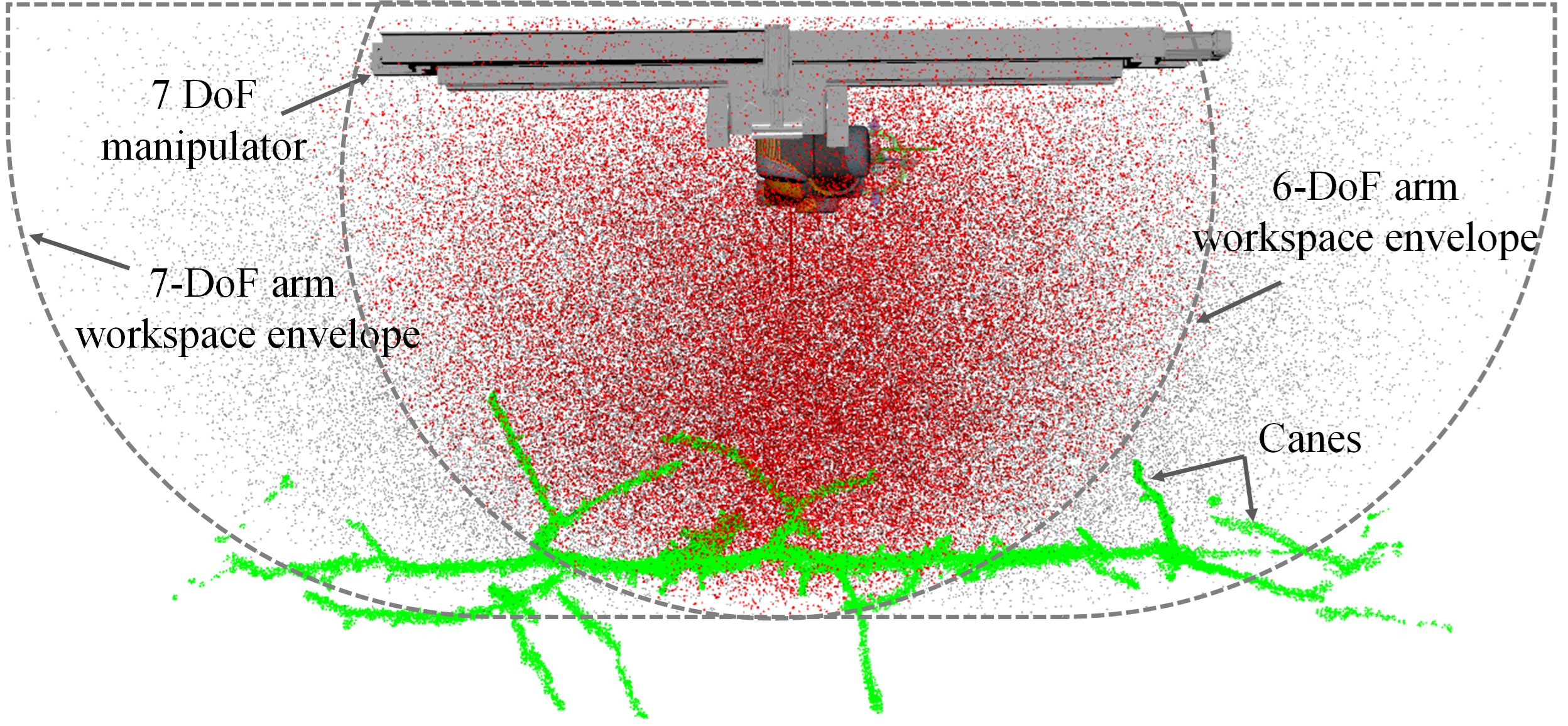}%
\label{fig:bumblebee-ws}}
\caption{(a) CAD rendering of the 7 DoF robot with its components. (b) Top view of the work volume of the 6 and 7 DoF arms.~\cite{silwal2022bumblebee}}
\label{fig:extended-DoF}
\end{figure*}

The Universal Robots UR5 manipulator \footnote{\url{https://universal-robots.com/products/ur5-robot/}}, which is a 6 DoF manipulator with 6 prismatic joints, has been successfully employed in several works. 
For instance, You et al.~\cite{you_efficient_2020} proved that it can be used to prune cherry trees when equipped with a cutter and an eye-in-hand Intel RealSense D435.  
A multi-pose approach was used to achieve the target cut points successfully. However, it should be noted that the study was carried out in a laboratory setting using a simplified tree environment, reducing the manipulation's complexity. The authors were clear that increasing the number of degrees of freedom would be key to improving timing within a sequence of goal poses. Furthermore, placing the robot on a gantry will effectively resolve the visual serving issue.
Botterill et al.~\cite{botterill_robot_2017} employed the same manipulator with 6 joints for pruning vine plants. The robot arm is considered the main cause of slow path execution times. This is because the arm sometimes requires a wide joint rotation to move the cutting tool over a short distance. Furthermore, the kinematics of a 6 DoF arm can only assume two poses (elbow-up and elbow-down), making collision avoidance extremely challenging. They also emphasized that an arm with more degrees of freedom or a different joint configuration will enable fast movements.
Furthermore, Silwal et al.~\cite{silwal2022bumblebee} demonstrate the limitations of the same 6 DoF robotic arm, which are determined by several factors, including singularities, self-collision, and collision models of the environment. They added a prismatic joint to the base to overcome the abovementioned limitations, obtaining a kinematically redundant design. This allows the end-effector to achieve any combination of orientation required to reach pruning locations. Furthermore, adding a degree of freedom increases the possibility of finding a solution for the motion of the arm. Adding the prismatic joint also increases the workspace, allowing the system to reach the whole vine plant from a single stationary point without moving the base. The result, showing the design of the 7 DoF manipulator and the improved workspace is shown in Fig.~\ref{fig:extended-DoF}

Teng et al.~\cite{teng_whole-body_2021} employed another commercial manipulator in their study, using a velocity-controlled two-wheel non-holonomic mobile robot, MP-500 (Neobotix GmbH. Co.)\footnote{\url{https://neobotix-roboter.de/}}, and a 7-DoFs robot arm (Franka Emika. Co.)\footnote{\url{https://franka.de/}}. Each had its own controller. The base localization algorithm employed odometry and twist information of the base central frame. The 7 DoF arm is composed of seven revolute joints and is analyzed in the paper, including its kinematics and dynamics.

In other works, the manipulator is designed directly, such as in Zhang et al.~\cite{zhang2022design}, where a 5-DoF arm is designed by studying the movement of the human body when pruning. They used this knowledge to create a machine that can perform the same task. The machine is composed of three main parts: a foundation support with a rotary joint, the machine body with a prismatic joint, and the large arm, which includes the elbow joint, the rotary joint of the forearm, and the forearm.  The link size parameter for each joint was selected based on the dimensions of the jujube tree before and after pruning, as well as the distance between the jujube trees in the row. Furthermore, we have created a three-dimensional mathematical simulation model of the manipulator using the MATLAB Robotics Toolbox. The manipulator workspace was calculated using the Monte Carlo Method, which yielded a range of -600 to -200 mm in the X, Y, and Z directions. In this case, the pruning points of the jujube were identified manually by an operator, and the testing trajectories were a reproduction of previously identified trajectories. The positioning errors of the end-effector of the pruning manipulator at different heights and depths are all less than 10 mm, as expected.
Zahid et al.~\cite{zahid2020development} focused their research on developing a 3 DoF end effector, which led to the design of a 3-prismatic DoF Cartesian manipulator. The placement of the first three prismatic joints is decided to move along the x-axis, y-axis, and z-axis, respectively. This is achieved using three linear actuators and a rigid base platform, which inevitably leads to vibration of the end effector. Therefore, a square base platform was placed to dampen vibrations and improve stability. Furthermore, a linear rigid arm was attached to the z-axis, enabling the Cartesian manipulator to be placed outside the tree canopy without interfering with the branches.

As an alternative, Lu~\cite{lu_kinematics_2021} proposed a dual-arm pruning robot design. In this work, the structure and the 3D model of the pruning robot are designed, including the building of the Denavit-Hartenberg kinematic model. Moreover, the workspace is analyzed, as well as the trajectory planning in the joint space. However, this research is at a beginning stage, and no in-field tests have been carried out.

\begin{table*}[!t]
\centering

\caption{The different configurations of pruning manipulators and relative performances.}
\begin{tabularx}{\textwidth}{l l p{7cm} c c}
    \toprule
    Crop & Configuration  & Performance & Year & Refe.\\ 
    \midrule
    Grapevine & 6 DoF: 6R & Cut success rate: 66\%, cut time 12s per vine plant & 2017 &~\cite{botterill_robot_2017}\\ \midrule[0.1pt]
    Cherry & 6 DoF 6R & Cut success rate: 75\%, cut time: 13 s, laboratory setup.& 2020 &~\cite{you_efficient_2020}\\ \midrule[0.1pt]
    Apple & 6 DoF: 3P + 3R & Cut success rate: 65\%, & 2020 &~\cite{zahid2020development}\\ \midrule[0.1pt]
    Grapevine & 7 DoF: 7R & Cut time: 28 s, laboratory setup. The time includes the approach of a non-holonomic robot & 2021 &~\cite{teng_whole-body_2021}\\ \midrule[0.1pt]
    - & 2 arms, 6 DoF: 6R & The design has been tested only in simulation. & 2021 &~\cite{lu_kinematics_2021} \\ \midrule[0.1pt]
    Grapevine & 7 DoF: 1P + 6R & Pruning accuracy: 87\%, cut time: 213 s per vine plant & 2022 &~\cite{silwal2022bumblebee}\\ \midrule[0.1pt]
    Jujube & 5 DoF: 4R, 1P & Max position error: 10 mm, success rate: 85.16\%, cut time: 29.3 min per Jujubee tree & 2022 &~\cite{zhang2022design} \\
    \bottomrule
\end{tabularx}
\label{tab:manipulator}
\end{table*}

\subsection{End-effector Design}
\label{subsec:end-eff}
Selecting the appropriate end effector is a crucial step in designing an autonomous pruning tool. Winter fruit tree branches are often complex, densely packed, and overlapping, presenting a significant challenge for a tool that must navigate freely to reach targeted areas. The design process must address both mechanical and spatial factors, such as size, shape, weight, and maneuverability, alongside the specific requirements of the branches themselves, considering their physical structure, horticultural characteristics, and biological properties~\cite{kondo1998robotics}.
The most common pruning techniques for fruit trees are shear cutting and saw cutting. Shear pruning is a supported method that provides a stable operation process, resulting in a clean and precise cut. The mechanism slices through branches, rather than tearing them. Saw cutting is better for thicker branches, but it produces a rougher cut surface. A complete overview of the methods proposed in the literature is summarized in Table~\ref{tab:endeffector}.

Given the structural characteristics of the articulated manipulator, Zhang et al.~\cite{zhang2022design} chose a shear mechanism as the end-effector for the jujube pruning manipulator. This mechanism consists of a drive motor, planetary reducer, gear transmission system, moving cutter, stationary cutter, diagonal photoelectric sensor, mounting plate, and fixed support. The robotic arm moves the end-effector mounted on the forearm to the target branch. Once the diagonal photoelectric sensor detects the branch in the scissor opening, the executive motor activates, closing the moving cutter. The jujube branches had a diameter of 5–20 mm to ensure efficient entry into the cutting mouth of the end-effector. The opening angle between the moving and fixed cutters was set at 40 degrees, with a maximum vertical distance of 35 mm at the scissor's mouth. Additionally, the distance from the cutting position of the jujube branch to the moving cutter's rotational axis was 50 mm.

To cut fruit trees such as cherries, You et al. developed a pneumatically-actuated four-bar linkage with custom ground blades, as detailed in their paper~\cite{you_efficient_2020}. In initial tests, the cutter proved capable of consistently cutting branches up to 10 mm in diameter near the pivot point of the blades. In a later work~\cite{you2023semiautonomous}, they employed an end-effector made up of an integrated battery-operated electric bypass pruner, which can be controlled via serial communication, and an Intel RealSense D435 RGB-D camera. The camera is positioned in an eye-in-hand configuration above the cutter and angled down 10 degrees, ensuring the top blade is always visible when the pruners are open. The selected shears are designed to cut branches with a diameter up to 3.2 cm. 

Some considerations about how the design of the end effector must take into account the variability of the mechanical properties of the canes in vine plants were ventilated by Silwal et al.~\cite{silwal2022bumblebee}. In fact, vines exhibit a wide variation in the length and diameter of dormant canes depending on nutrients and water. Furthermore, dead samples of vines required a higher force to cut. The end-effector designed in this work fixes one of the handles to a rigid surface, while the weight needed to cut canes of different diameters was evaluated by applying different weights to the other handle. The movable scissor is actuated by a combination of high-torque and floating pulleys, taking into account the robot's payload capacity.

To address the problem of obtaining all the degrees of freedom needed for robotic pruning, Zahid et al.~\cite{zahid2020development} designed a 3 revolute DoF end effector, as the continuation of a Cartesian 3P manipulator. The primary design criteria must include maneuverability and spatial requirements to precisely position the cutter at a specific orientation, minimizing space utilization. The designed model consists of three motors with their rotational axes perpendicular to each other. Finally, a shear cutter was integrated with the end-effector as a cutting tool to produce smooth pruning cuts. The diameter of apple branches was considered, which is usually less than 25 mm, and a front opening of 60 mm was used.

The problem of metal support wires was solved by Williams et al.~\cite{williams2024archie}, who designed a custom cutting system, called Barracuda, based on small recesses on the bigger teeth of the cutter. It allows the wires not to be cut and to be pushed into the small recesses between teeth, while the thicker cane, which does not fit into the narrow gaps, is cut by the blades.

As alternative, saw cutting was explored by Botterill et al.~\cite{botterill_robot_2017}, where a 6 mm CNC router mill-end attached to a 100 W, 24k rpm brushless DC motor was employed. To cut a cane, the robot arm sweeps the cutter through the cane. However, as the cut motion requires a considerable collision-free space, the cutter sometimes fails to make separation cuts. The authors have suggested adopting a secateur-based cutter in the future.

\begin{table*}
\centering

\caption{The different designs of the end-effectors and the respective performances.}
\begin{tabularx}{\textwidth}{lXXlc}
    \toprule
    Crop & Mechanism and actuation  & Performances & Year & Ref.\\ 
    \midrule
    Apple & Shear, electromechanical &- & 2020 &~\cite{zahid2020development}\\ \midrule[0.1pt]
    Cherry & Shear, pneumatic & 92\% of cutting success & 2020 &~\cite{you_efficient_2020}\\ \midrule[0.1pt]
    Grape & Saw (mill), electromechanical & 66\% of cutting success & 2020 &~\cite{zahid2020development}\\ \midrule[0.1pt]
    Jujubee & Shear, electromechanical   & Max cutting diameter: 12 mm, time per cut: 1 s  & 2022 &~\cite{zhang2022design}\\ \midrule[0.1pt]
    Grape & Shear, electromechanical &- & 2022 &~\cite{silwal2022bumblebee}\\ \midrule[0.1pt]
    Cherry & Shear, electromechanical &- & 2023 &~\cite{you2023semiautonomous}\\ \midrule[0.1pt]
    Grape & Shears with gaps for wires, electromechanical & 97.0\% cutting success & 2024 &~\cite{williams2024archie} \\
    
    \bottomrule
\end{tabularx}
\label{tab:endeffector}

\end{table*}

\subsection{Path planning and control}
\label{subsec:planning_control}

Once a pruning point is detected and localized, the final step involves planning the trajectory of the end effector to reach the target and execute a precise cut. When the estimate is accurate, the key challenge becomes computing a collision-free path in the presence of environmental constraints.cTeng et al.~\cite{teng_whole-body_2021} proposed a hierarchical control strategy for highly redundant robots to manage complex tasks in Cartesian space. Their system exploits self-motion redundancy to prioritize primary tasks while accommodating secondary constraints. A task-space segmentation method enables concurrent handling of both, and joint velocities are computed via an incremental Jacobian matrix~\cite{slotine1991general}. A visual perception system~\cite{fernandes_grapevine_2021} identifies pruning points, followed by quintic polynomial interpolation~\cite{guan2005robotic} for smooth trajectory generation. The robot’s non-holonomic base helps navigate joint limits and singularities.

\begin{figure}
    \centering
    \includegraphics[width=0.75\linewidth]{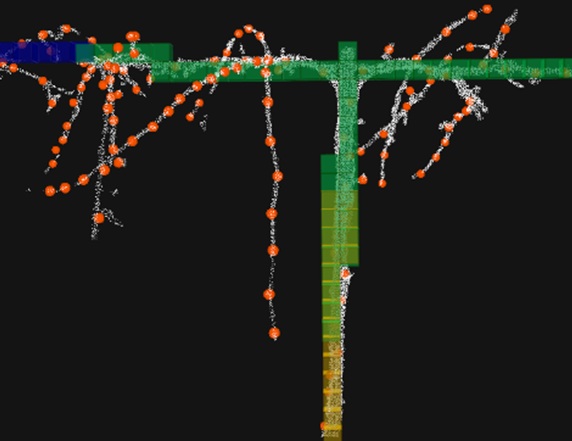}
    \caption{Cordon and trellis wire are considered hard obstacles for motion planning~\cite{silwal2022bumblebee}.}
    \label{fig:obstacle-avoidance}
\end{figure}

The Rapidly-exploring Random Tree (RRT)-Connect algorithm~\cite{kuffner2000rrt} is widely used for motion planning in unstructured environments. Chen et al.~\cite{chen2022path} introduced an improved version with goal-biased sampling and adaptive step-size adjustment, paired with cubic B-spline interpolation to smooth paths. Their method reduced path length and planning time by $60\%$ and $55\%$, respectively, achieving full obstacle avoidance. Similarly, You et al.~\cite{you2023semiautonomous} used RRT-Connect in a stop-and-go approach, where a mobile base brings the robot near pruning points, reducing large joint motions. A hybrid visual-force controller guides alignment and contact, and electric cutters ensure minimal disturbance during cutting. Silwal et al.~\cite{silwal2017design} adopted a 7-DoF RRT-Connect planner, outperforming alternatives in prior studies~\cite{paulin2015comparison}. Their method divides the trajectory into two phases: RRT-Connect brings the tool near the target, followed by Cartesian planning for final alignment. A retract-and-reset routine improves collision avoidance and path quality, considering obstacles as in Fig.~\ref{fig:obstacle-avoidance}. Botterill et al.~\cite{botterill_robot_2017} also used RRT-Connect with goal-biased sampling and geometric primitives for collision detection, dynamically adjusting safety margins to streamline path planning.

You et al.~\cite{you_efficient_2020} introduced a database-driven method, discretizing the workspace into a grid of optimized poses for rapid online planning. Trajectories are sequenced by solving a Traveling Salesman Problem (TSP), leveraging methods like CHOMP~\cite{zucker2013chomp} and STOMP~\cite{schulman2013finding}. Proportional control handles linear motions, with BIT*~\cite{gammell2015batch} used for replanning when needed.

Recent work has explored model-free control to improve adaptability. While model-based approaches require accurate task models, Reinforcement Learning (RL) enables agents to learn control policies through trial-and-error in simulation. You et al.~\cite{you2022precision} incorporated an RL-based visual controller that identifies cut points from camera input. Rewards encourage branch visibility and cutter alignment, while a force-based admittance controller finalizes the cut. This hybrid strategy improves performance in dynamic environments by combining learned and reactive control.

Learning from demonstration represents a promising alternative, enabling robots to imitate human pruning strategies without relying on reward signals. Advances in simulation and VR further support this approach, offering a natural and intuitive path for training pruning behaviors.

\section{Complete Robotic Pruning Systems}
\label{sec:complete}

\begin{figure}[!t]
\centering
\subfloat[]{\includegraphics[width=0.48\textwidth]{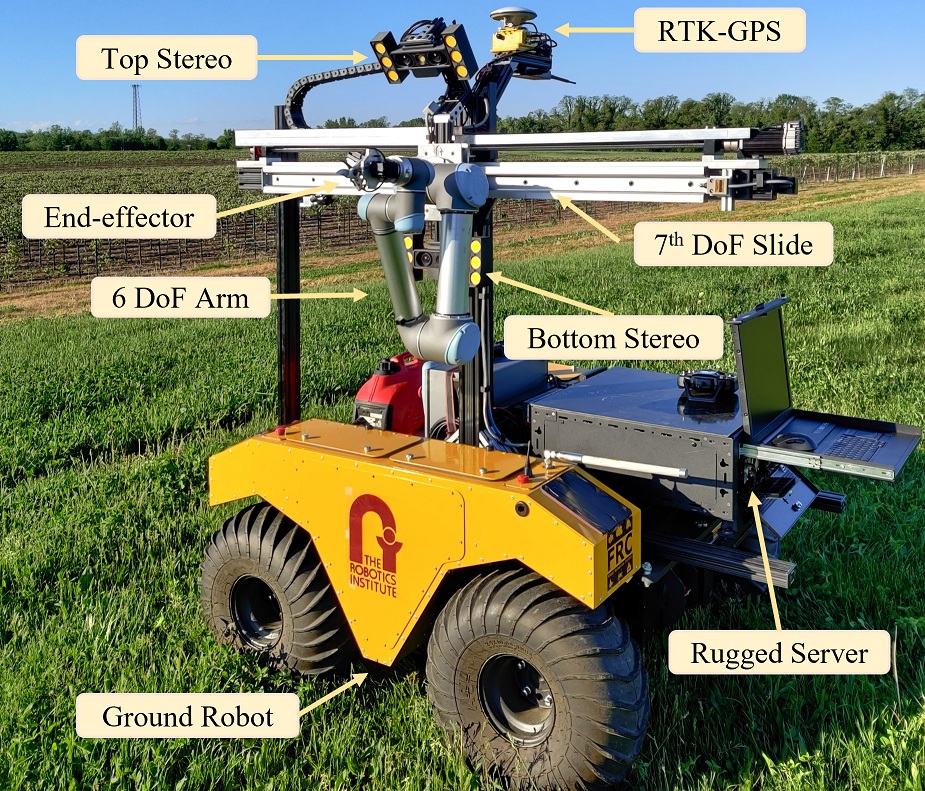}%
\label{fig:bumblebee-full}}
\hfil
\subfloat[]{\includegraphics[width=0.48\textwidth]{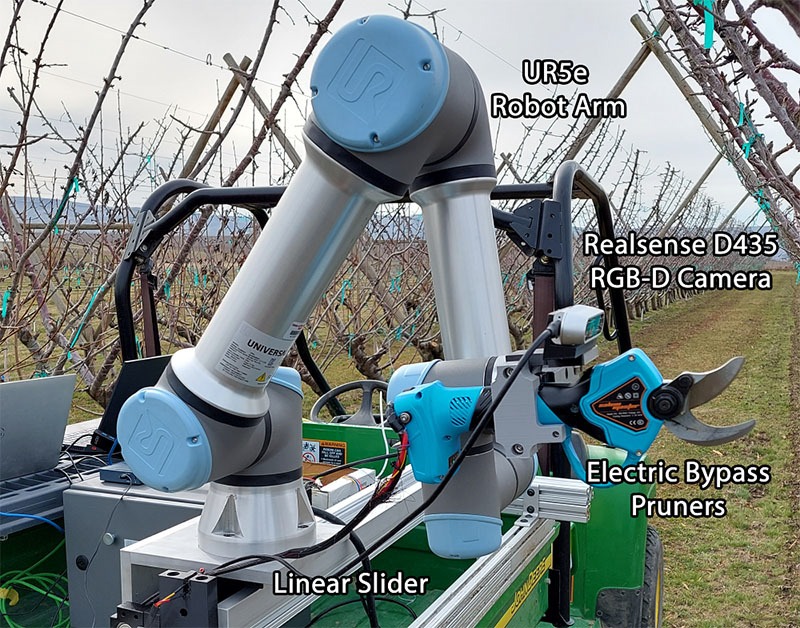}%
\label{fig:you-full}}
\caption{Integrated robotic systems: (a): the system comprises a rover, a 7 DoF robot arm, cutting end-effector, dual stereo cameras and on-board computers~\cite{silwal2022bumblebee}; (b):  the system comprises a 6 DoF robot arm, a cutting end-effector, a stereo camera and on-board computer~\cite{you_autonomous_2022}.}
\label{fig:full-systems}
\end{figure}

Several studies have proposed comprehensive end-to-end pruning pipelines that integrate perception, control, and the design of both the robot and its end-effector. These integrated systems are designed to operate autonomously and efficiently across diverse environments. The perception modules typically employ advanced sensors and algorithms to detect and analyze environmental features, allowing for accurate target identification. Control systems rely on sophisticated motion planning and execution software to ensure precise and coordinated movements. Meanwhile, the robot and end-effector designs are tailored specifically for pruning tasks, incorporating characteristics like dexterity, reach, and durability. These complete systems are validated through extensive field trials to ensure reliability and effectiveness under real-world conditions.

Silwal et al.~\cite{silwal2022bumblebee} introduced a fully autonomous system for dormant season grapevine pruning, as shown in Fig.~\ref{fig:bumblebee-full}. Their solution combines a vision system, a ground robot, a kinematically redundant 7-DoF manipulator—achieved by adding a prismatic joint to a standard 6-DoF UR5—and purpose-built pruning algorithms. The end-effector was specifically designed to handle the cutting forces correlated with branch diameter. To reduce the impact of uneven illumination, the robot employs an active light stereo camera system~\cite{silwal2021robust} that reconstructs dual point clouds. These are registered using transformation matrices and refined with the ICP algorithm. Bud detection—a critical step in grapevine pruning—is carried out with a Faster R-CNN model. To avoid collisions with trellis structures and wires, a RANSAC algorithm identifies vertical and horizontal line constraints in the 3D vine model. In contrast, Corbett-Davies et al.~\cite{corbett-davies_expert_2012} focus on the decision-making logic for pruning, rather than hardware. They developed an AI-based system trained to replicate expert decisions using simulated 3D vine structures. Comparative assessments showed that their system outperformed a typical human pruner in 30

Botterill et al.~\cite{botterill_robot_2017} presented a mobile robotic platform that straddles vine rows and captures images using trinocular stereo cameras. A 3D model of each vine is constructed via feature matching, triangulation, and incremental bundle adjustment. A 6-DoF robotic arm, guided by an AI system, executes pruning actions based on the reconstructed structure. The robot arm's trajectory is planned in real-time using a rapidly-exploring random tree algorithm, achieving collision-free paths within 1.5 seconds per vine. Although pruning each vine takes around 2 minutes—comparable to human performance—system reliability is limited by the interdependencies of perception, planning, and actuation components.

In a different crop domain, You et al.~\cite{you_autonomous_2022} designed a pruning system for sweet cherry trees grown under the upright fruiting offshoot (UFO) architecture as shown in Fig.~\ref{fig:you-full}. Their fully autonomous platform integrates prior research in perception and manipulation, achieving a 58\% cutting success rate during field trials. While not yet robust enough for commercial deployment, this is a pioneering effort in robotic pruning for fruit trees and sets a solid groundwork for future improvements. To address the reality gap between simulation and field conditions, You et al.~\cite{you2022precision} also developed a simulated pruning environment featuring planar cherry trees and pre-labeled cutting points. Due to the noise in depth data, they opted for a 2D image-based control system. Using a 6-DoF manipulator with an Intel D435 camera, they trained a Generative Adversarial Network (GAN) to segment branches, enabling the visual controller to function reliably across simulated and real images. An actor-critic reinforcement learning algorithm then generated control signals for branch alignment. Upon contact, control transitions to an interaction-based controller that adjusts the cutter’s pivot to minimize forces during cutting.

\begin{figure}[!t]
\centering
\includegraphics[width=0.75\linewidth]{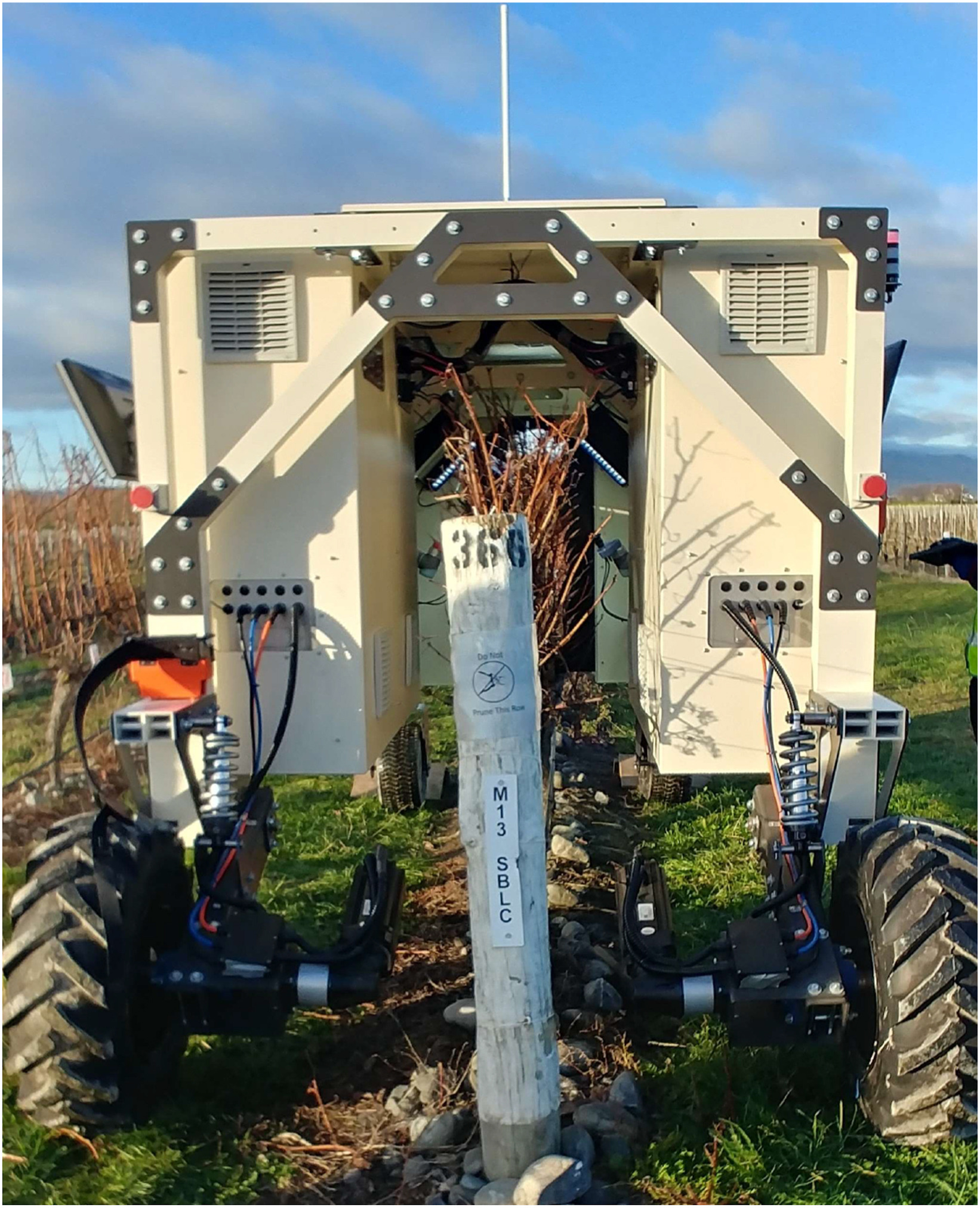}
\caption{Archie Jnr robotic platform, proposed by~\cite{williams2023modelling}.}
\label{fig:williams-full}
\end{figure}

Lastly, Williams et al.~\cite{williams2024archie} developed Archie Jnr, a fully operational pruning platform designed for a controlled environment under an arched structure (see Fig.~\ref{fig:williams-full}). The system uses two UR5 robotic arms to scan plants and conduct panoptic segmentation of nodes and canes for precise 3D reconstruction. A Charuco board is employed for accurate image-depth registration. The manipulators incorporate custom mechanisms that enhance agility, making this system especially suitable for precision pruning in high-control settings.

Collectively, these studies highlight the critical role of system integration in achieving autonomous pruning. Robust perception methods—particularly 3D reconstruction and deep learning-based segmentation—enable accurate identification of pruning targets. Control strategies range from traditional trajectory planning to learning-based policies and hybrid force-vision controllers. Manipulator designs are adapted to the physical demands of the pruning task, while platform-level decisions reflect a balance between robustness in the field and operational simplification in controlled environments.

\section{Discussion: challenges, solutions and future scenarios}
\label{sec:discussion}





\subsection{Challenges and Limitations of State of the Art}
\label{subsec:challenges}

The previous chapters have provided a broad overview of the existing solutions for developing an autonomous pruning system. Given the diversity of tasks involved and the range of expertise required to develop each component of such a system, it has become evident that a modular approach to robot design is fundamental for achieving optimal results. As demonstrated throughout this paper, most existing works focus on narrow aspects of the problem, such as branch detection, skeletonization, pruning point estimation, or end-effector design. Each component demands highly specialized skills spanning computer vision, agronomy, mechanical engineering, and control systems. Moreover, a modular platform can be repurposed for various agricultural tasks, such as health monitoring, fruit counting, and harvesting. For instance, a module designed for branch detection can also be used for obstacle avoidance during harvesting. At the same time, the manipulator structure may remain the same, with only the end-effector swapped for the target task.

However, several limitations have emerged from the current state of the art. One of the most significant barriers to large-scale deployment remains the high cost of robotic platforms. Pruning, unlike many other agricultural operations, demands labor-intensive interventions on a per-plant basis, which, when scaled to large fields, results in substantial labor and operational costs. A major cost factor lies in the sensors themselves. While high-performance sensors are expensive, recent developments in AI and software optimization have shown that effective performance can be achieved using fewer and lower-cost sensors. Furthermore, software development represents a fixed cost during R\&D, as opposed to recurring hardware expenses incurred for each robot.

Another critical limitation is operational speed. Current systems require between one minute per branch to two minutes per vine plant (typically involving about eight cuts), which is an order of magnitude slower than manual pruning. As noted by~\cite{botterill_robot_2017}, one possible short-term solution is the adoption of faster robotic arms. However, addressing this issue more comprehensively requires improvements to the detection and control algorithms, which remain computationally demanding. Two primary strategies can be pursued:
First, optimization of the algorithms through lighter-weight neural networks and more efficient processing pipelines may yield similar performance with lower computational overhead. Second, more powerful on-board hardware can be adopted to meet the processing demands required for real-time inference and control, enabling higher actuation speeds without compromising accuracy.

Most of the proposed algorithms for branch detection, segmentation, 3D reconstruction, and pruning point estimation rely on learning-based methods. While this trend is globally evident, a significant drawback of such approaches is the requirement for large, labeled datasets for effective training. Constructing a structured and diverse dataset is labor-intensive, particularly due to the complexity of manually labeling pruning-relevant features. Several works have attempted to address this challenge by developing realistic simulation environments to generate synthetic training data. These environments can simulate plant structures and growth patterns based on agronomic models, allowing for the automatic generation of diverse and labeled training data. This approach not only facilitates training of detection and decision algorithms but also enables analysis of the short- and long-term consequences of different pruning strategies. Models trained in simulation and then fine-tuned with real-world data have shown comparable performance to those trained entirely on real datasets. This is a significant advantage in pruning applications, which are inherently destructive, making real-world testing both costly and risky in terms of yield loss.

A few studies offer complete end-to-end solutions that integrate all stages of the pruning pipeline—from perception and decision-making to robot and end-effector design. These systems are typically tested under real-world field conditions to assess performance and reliability. For instance,~\cite{you_autonomous_2022} presents an autonomous pruner for upright fruiting offshoo cherry trees, which integrates a semantic segmentation algorithm for plant structure identification, a pruning point estimator, and a commercial manipulator equipped with a custom electromechanical end-effector. Improved robustness is demonstrated in~\cite{you2022precision}, where the training process incorporates a simulation environment powered by Generative Adversarial Networks (GANs)~\cite{goodfellow2014generative}, following a sim-to-real approach similar to~\cite{james2019sim}. This reduces the gap between synthetic and real-world data.
Another successful example is the work of~\cite{silwal2022bumblebee}, which includes the complete design of both manipulator and end-effector, as well as integration of perception systems. In this case, using active lighting techniques~\cite{silwal2021robust} enhances the accuracy of bud detection and pruning point estimation. Active lighting also improves the quality of the 3D point cloud (PCL), enabling more accurate plant reconstruction and thus more reliable motion planning and control. The complete system is deployed on a mobile platform capable of autonomous navigation within the field using GPS with RTK corrections.

\subsection{Future Scenarios}
\label{subsec:future}
One promising direction involves developing autonomous pruning machines prioritizing robustness and precision over compactness. For instance, the system proposed in~\cite{botterill_robot_2017} operates in a partially enclosed, controlled environment, where uniform lighting and backgrounds enhance segmentation accuracy. This setup also supports the use of high-performance sensors, which can significantly improve depth perception—a key requirement for precise control of robotic arms and end-effectors. While lower-cost sensors are attractive from an economic standpoint, they may compromise accuracy, potentially leading to suboptimal cuts or even crop loss. Future research could focus on hybrid solutions that balance hardware cost with intelligent compensation through software.

Emerging machine learning paradigms offer new approaches that could bypass traditional step-wise pipelines. Instead of detecting, segmenting, and analyzing plant structure in sequence, it may be possible to learn a direct mapping from sensory input to pruning actions. Reinforcement Learning (RL), for instance, allows an agent to learn optimal pruning policies through trial and error in a simulation environment, where pruning consequences can be integrated into the reward function. While promising for structured crops like apples, mangoes, and cherries, RL alone may be insufficient for grapevines due to the complex, region-specific pruning rules they require.

To address domain-specific nuances, Learning from Demonstration (LfD) provides a powerful paradigm. By observing expert behavior, a pruning policy can be fine-tuned for both structural accuracy and agronomic validity. For instance,~\cite{lauretti2023robot} shows how demonstration-based approaches can avoid motion discontinuities in complex tasks. Complementary technologies like augmented and virtual reality could support expert-guided training, allowing humans to annotate or demonstrate pruning strategies in an immersive, simulated environment. An example is~\cite{haring2024vid2cuts}, where an AR mobile app provides pruning suggestions to non-experts using AI-generated recommendations.

Another promising avenue is imitation learning, which has succeeded in tasks like pepper harvesting~\cite{kim2025autonomous}. There, a visuomotor policy trained via demonstrations enabled a handheld shear-gripper to adapt to variable field conditions and crop types. Applying similar strategies to pruning could enhance adaptability and robustness, especially in unstructured outdoor environments. However, application in pruning remains rare, representing an untapped area for future investigation.

While the potential of AI-driven pruning is evident, future systems must integrate precision hardware, data-efficient learning, and domain-specific agronomic knowledge. Developing scalable, adaptable, and accurate solutions will likely require multi-disciplinary collaboration spanning robotics, machine learning, agronomy, and human-computer interaction.

\section{Conclusions}
\label{sec:conclusions}

This work has illustrated that autonomous pruning remains an open and highly relevant research challenge, demanding progress in robotics, perception, machine learning, and agronomic integration. While modern orchard training systems and agricultural machinery have streamlined many tasks, such as spraying, fertilization, and green pruning, dormant and selective pruning remain among the most labor-intensive and skill-dependent operations. These operations are crucial, as they directly affect the yield quality and quantity of the following season.

A core requirement for reliable autonomous pruning is the accurate understanding of plant structure, which is inherently unstructured, variable, and occluded. Traditional image processing approaches have shown limited performance in complex outdoor scenes. In contrast, deep learning-based methods, particularly convolutional neural networks, have significantly improved the performance of branch detection and segmentation. These networks allow not only for detecting branches but also for classifying them hierarchically, which is critical for subsequent 3D reconstruction and pruning logic.
Detecting buds is another essential capability, as pruning rules often depend on the position and condition of buds. However, their small size and inconsistent visibility make detection challenging, typically requiring high-resolution cameras and precise segmentation.
The next logical step after segmentation is skeletonization, which helps define the structural hierarchy and spatial configuration of the plant. This, in turn, supports pruning point estimation, for which two main methodological categories exist: rule-based methods that extract pruning points from a reconstructed structure using deterministic logic learning-based approaches that estimate pruning points directly from raw data.

Beyond perception, the success of autonomous pruning also hinges on the robot’s physical hardware, including the manipulator and end-effector. While perception determines where to prune, the robot must also be capable of acting safely and precisely. Recent studies have developed custom electromechanical end-effectors capable of adapting to varying branch sizes and minimizing damage. Similarly, the choice of manipulator, such as its reach, degrees of freedom, and control workspace, directly affects system flexibility and responsiveness in field conditions. Speed remains a bottleneck; current systems are often an order of magnitude slower than manual pruning, highlighting the need for faster arms and more efficient planning algorithms.

Finally, the integration of all these components—perception, planning, actuation, and platform mobility—into fully functional, field-ready pruning robots is still in early stages. Some examples demonstrate the feasibility of such systems in specific scenarios, combining semantic segmentation, GPS navigation, and simulation-based training. These systems highlight both the potential and the complexity of real-world deployment. Notably, the modular design of these robots allows them to be adapted for other agricultural tasks, such as health monitoring or harvesting, thereby improving their return on investment.

In summary, although many technical components have shown promise in isolation, fully integrated, generalizable autonomous pruning systems require further development. Future research must focus on bridging the gap between lab prototypes and robust, scalable platforms, emphasizing generalization, simulation-based training, and human-in-the-loop learning methods. Ultimately, the path forward lies in a multidisciplinary collaboration that combines agronomic expertise with cutting-edge robotics and AI, unlocking a new era of intelligent, sustainable agriculture.

\bibliographystyle{elsarticle-num} 
\bibliography{biblio}

\begin{thebibliography}{100}
\expandafter\ifx\csname url\endcsname\relax
  \def\url#1{\texttt{#1}}\fi
\expandafter\ifx\csname urlprefix\endcsname\relax\def\urlprefix{URL }\fi
\expandafter\ifx\csname href\endcsname\relax
  \def\href#1#2{#2} \def\path#1{#1}\fi

\bibitem{galinato2022costs}
S.~P. Galinato, A.~Kendall, C.~A. Miles, Costs and profitability for mechanized pruning and harvest in two cider apple orchard systems, HortTechnology 32~(3) (2022) 275--287.

\bibitem{europe2024fruit}
{European Commission}, The fruit and vegetable sector in the eu - a statistical overview, Eurostat (2024).

\bibitem{europe2024farmers}
{European Commission}, Farmers and the agricultural labour force - statistics, Eurostat (2022).

\bibitem{calvin2022adjusting}
L.~Calvin, P.~Martin, S.~Simnitt, Adjusting to higher labor costs in selected us fresh fruit and vegetable industries (2022).

\bibitem{allegro2022effects}
G.~Allegro, R.~Martelli, G.~Valentini, C.~Pastore, R.~Mazzoleni, F.~Pezzi, I.~Filippetti, Effects of mechanical winter pruning on vine performances and management costs in a trebbiano romagnolo vineyard: A five-year study, Horticulturae 9~(1) (2022) 21.

\bibitem{he_sensing_2018}
L.~He, J.~Schupp, \href{http://www.mdpi.com/2073-4395/8/10/211}{Sensing and {Automation} in {Pruning} of {Apple} {Trees}: {A} {Review}}, Agronomy 8~(10) (2018) 211.
\newblock \href {https://doi.org/10.3390/agronomy8100211} {\path{doi:10.3390/agronomy8100211}}.
\newline\urlprefix\url{http://www.mdpi.com/2073-4395/8/10/211}

\bibitem{tinoco2021review}
V.~Tinoco, M.~F. Silva, F.~N. Santos, L.~F. Rocha, S.~Magalh{\~a}es, L.~C. Santos, A review of pruning and harvesting manipulators, in: 2021 IEEE International Conference on Autonomous Robot Systems and Competitions (ICARSC), IEEE, 2021, pp. 155--160.

\bibitem{zahid2021technological}
A.~Zahid, M.~S. Mahmud, L.~He, P.~Heinemann, D.~Choi, J.~Schupp, Technological advancements towards developing a robotic pruner for apple trees: A review, Computers and Electronics in Agriculture 189 (2021) 106383.

\bibitem{qiang2014identification}
L.~Qiang, C.~Jianrong, L.~Bin, D.~Lie, Z.~Yajing, Identification of fruit and branch in natural scenes for citrus harvesting robot using machine vision and support vector machine, International Journal of Agricultural and Biological Engineering 7~(2) (2014) 115--121.

\bibitem{ji2016apple}
W.~Ji, Z.~Qian, B.~Xu, Y.~Tao, D.~Zhao, S.~Ding, Apple tree branch segmentation from images with small gray-level difference for agricultural harvesting robot, Optik 127~(23) (2016) 11173--11182.

\bibitem{shalal2015orchard}
N.~Shalal, T.~Low, C.~McCarthy, N.~Hancock, Orchard mapping and mobile robot localisation using on-board camera and laser scanner data fusion--part a: Tree detection, Computers and Electronics in Agriculture 119 (2015) 254--266.

\bibitem{gao2006image}
M.~Gao, T.-F. Lu, Image processing and analysis for autonomous grapevine pruning, in: 2006 international conference on mechatronics and automation, IEEE, 2006, pp. 922--927.

\bibitem{mcfarlane1997image}
N.~McFarlane, B.~Tisseyre, C.~Sinfort, R.~Tillett, F.~Sevila, Image analysis for pruning of long wood grape vines, Journal of agricultural engineering research 66~(2) (1997) 111--119.

\bibitem{zhang2018branch}
J.~Zhang, L.~He, M.~Karkee, Q.~Zhang, X.~Zhang, Z.~Gao, Branch detection for apple trees trained in fruiting wall architecture using depth features and regions-convolutional neural network (r-cnn), Computers and Electronics in Agriculture 155 (2018) 386--393.

\bibitem{amatya2017automated}
S.~Amatya, M.~Karkee, Q.~Zhang, M.~D. Whiting, \href{http://www.mdpi.com/2218-6581/6/4/31}{Automated {Detection} of {Branch} {Shaking} {Locations} for {Robotic} {Cherry} {Harvesting} {Using} {Machine} {Vision}}, Robotics 6~(4) (2017) 31.
\newblock \href {https://doi.org/10.3390/robotics6040031} {\path{doi:10.3390/robotics6040031}}.
\newline\urlprefix\url{http://www.mdpi.com/2218-6581/6/4/31}

\bibitem{amatya2016integration}
S.~Amatya, M.~Karkee, Integration of visible branch sections and cherry clusters for detecting cherry tree branches in dense foliage canopies, Biosystems Engineering 149 (2016) 72--81.

\bibitem{botterill_robot_2017}
T.~Botterill, S.~Paulin, R.~Green, S.~Williams, J.~Lin, V.~Saxton, S.~Mills, X.~Chen, S.~Corbett-Davies, \href{https://onlinelibrary.wiley.com/doi/abs/10.1002/rob.21680}{A {Robot} {System} for {Pruning} {Grape} {Vines}}, Journal of Field Robotics 34~(6) (2017) 1100--1122, \_eprint: https://onlinelibrary.wiley.com/doi/pdf/10.1002/rob.21680.
\newblock \href {https://doi.org/10.1002/rob.21680} {\path{doi:10.1002/rob.21680}}.
\newline\urlprefix\url{https://onlinelibrary.wiley.com/doi/abs/10.1002/rob.21680}

\bibitem{you2022semantics}
A.~You, C.~Grimm, A.~Silwal, J.~R. Davidson, Semantics-guided skeletonization of upright fruiting offshoot trees for robotic pruning, Computers and Electronics in Agriculture 192 (2022) 106622.

\bibitem{oliveira2024enhancing}
F.~Oliveira, D.~Q. da~Silva, V.~Filipe, T.~M. Pinho, M.~Cunha, J.~B. Cunha, F.~N. Dos~Santos, Enhancing grapevine node detection to support pruning automation: Leveraging state-of-the-art yolo detection models for 2d image analysis, Sensors 24~(21) (2024) 6774.

\bibitem{williams2023modelling}
H.~Williams, D.~Smith, J.~Shahabi, T.~Gee, M.~Nejati, B.~McGuinness, K.~Black, J.~Tobias, R.~Jangali, H.~Lim, et~al., Modelling wine grapevines for autonomous robotic cane pruning, biosystems engineering 235 (2023) 31--49.

\bibitem{majeed_apple_2018}
Y.~Majeed, J.~Zhang, X.~Zhang, L.~Fu, M.~Karkee, Q.~Zhang, M.~D. Whiting, \href{https://linkinghub.elsevier.com/retrieve/pii/S2405896318311807}{Apple {Tree} {Trunk} and {Branch} {Segmentation} for {Automatic} {Trellis} {Training} {Using} {Convolutional} {Neural} {Network} {Based} {Semantic} {Segmentation}}, IFAC-PapersOnLine 51~(17) (2018) 75--80.
\newblock \href {https://doi.org/10.1016/j.ifacol.2018.08.064} {\path{doi:10.1016/j.ifacol.2018.08.064}}.
\newline\urlprefix\url{https://linkinghub.elsevier.com/retrieve/pii/S2405896318311807}

\bibitem{majeed2019study}
Y.~Majeed, M.~Karkee, Q.~Zhang, L.~Fu, M.~D. Whiting, \href{https://www.sciencedirect.com/science/article/pii/S2405896319324176}{A study on the detection of visible parts of cordons using deep learning networks for automated green shoot thinning in vineyards}, IFAC-PapersOnLine 52~(30) (2019) 82--86, 6th IFAC Conference on Sensing, Control and Automation Technologies for Agriculture AGRICONTROL 2019.
\newblock \href {https://doi.org/https://doi.org/10.1016/j.ifacol.2019.12.501} {\path{doi:https://doi.org/10.1016/j.ifacol.2019.12.501}}.
\newline\urlprefix\url{https://www.sciencedirect.com/science/article/pii/S2405896319324176}

\bibitem{wu_extracting_2020}
J.~Wu, G.~Yang, H.~Yang, Y.~Zhu, Z.~Li, L.~Lei, C.~Zhao, \href{https://linkinghub.elsevier.com/retrieve/pii/S0168169920301605}{Extracting apple tree crown information from remote imagery using deep learning}, Computers and Electronics in Agriculture 174 (2020) 105504.
\newblock \href {https://doi.org/10.1016/j.compag.2020.105504} {\path{doi:10.1016/j.compag.2020.105504}}.
\newline\urlprefix\url{https://linkinghub.elsevier.com/retrieve/pii/S0168169920301605}

\bibitem{majeed2020deep}
Y.~Majeed, J.~Zhang, X.~Zhang, L.~Fu, M.~Karkee, Q.~Zhang, M.~D. Whiting, Deep learning based segmentation for automated training of apple trees on trellis wires, Computers and Electronics in Agriculture 170 (2020) 105277.

\bibitem{majeed2020determining}
Y.~Majeed, M.~Karkee, Q.~Zhang, L.~Fu, M.~D. Whiting, Determining grapevine cordon shape for automated green shoot thinning using semantic segmentation-based deep learning networks, Computers and Electronics in Agriculture 171 (2020) 105308.

\bibitem{fernandes_grapevine_2021}
M.~Fernandes, A.~Scaldaferri, G.~Fiameni, T.~Teng, M.~Gatti, S.~Poni, C.~Semini, D.~Caldwell, F.~Chen, Grapevine {Winter} {Pruning} {Automation}: {On} {Potential} {Pruning} {Points} {Detection} through {2D} {Plant} {Modeling} using {Grapevine} {Segmentation}, in: 2021 {IEEE} 11th {Annual} {International} {Conference} on {CYBER} {Technology} in {Automation}, {Control}, and {Intelligent} {Systems} ({CYBER}), 2021, pp. 13--18, iSSN: 2642-6633.
\newblock \href {https://doi.org/10.1109/CYBER53097.2021.9588303} {\path{doi:10.1109/CYBER53097.2021.9588303}}.

\bibitem{lin_three-dimensional_2021}
G.~Lin, Y.~Tang, X.~Zou, C.~Wang, \href{https://linkinghub.elsevier.com/retrieve/pii/S0168169921001253}{Three-dimensional reconstruction of guava fruits and branches using instance segmentation and geometry analysis}, Computers and Electronics in Agriculture 184 (2021) 106107.
\newblock \href {https://doi.org/10.1016/j.compag.2021.106107} {\path{doi:10.1016/j.compag.2021.106107}}.
\newline\urlprefix\url{https://linkinghub.elsevier.com/retrieve/pii/S0168169921001253}

\bibitem{ma_automatic_2021}
B.~Ma, J.~Du, L.~Wang, H.~Jiang, M.~Zhou, \href{https://linkinghub.elsevier.com/retrieve/pii/S0168169921005019}{Automatic branch detection of jujube trees based on {3D} reconstruction for dormant pruning using the deep learning-based method}, Computers and Electronics in Agriculture 190 (2021) 106484.
\newblock \href {https://doi.org/10.1016/j.compag.2021.106484} {\path{doi:10.1016/j.compag.2021.106484}}.
\newline\urlprefix\url{https://linkinghub.elsevier.com/retrieve/pii/S0168169921005019}

\bibitem{tong_branch_2022}
S.~Tong, Y.~Yue, W.~Li, Y.~Wang, F.~Kang, C.~Feng, \href{https://www.mdpi.com/2072-4292/14/18/4495}{Branch {Identification} and {Junction} {Points} {Location} for {Apple} {Trees} {Based} on {Deep} {Learning}}, Remote Sensing 14~(18) (2022) 4495.
\newblock \href {https://doi.org/10.3390/rs14184495} {\path{doi:10.3390/rs14184495}}.
\newline\urlprefix\url{https://www.mdpi.com/2072-4292/14/18/4495}

\bibitem{you2022optical}
A.~You, C.~Grimm, J.~R. Davidson, Optical flow-based branch segmentation for complex orchard environments, in: 2022 IEEE/RSJ International Conference on Intelligent Robots and Systems (IROS), IEEE, 2022, pp. 9180--9186.

\bibitem{you2022precision}
A.~You, H.~Kolano, N.~Parayil, C.~Grimm, J.~R. Davidson, Precision fruit tree pruning using a learned hybrid vision/interaction controller, in: 2022 International Conference on Robotics and Automation (ICRA), IEEE, 2022, pp. 2280--2286.

\bibitem{borrenpohl2023automated}
D.~Borrenpohl, M.~Karkee, Automated pruning decisions in dormant sweet cherry canopies using instance segmentation, Computers and Electronics in Agriculture 207 (2023) 107716.

\bibitem{guadagna2023using}
P.~Guadagna, M.~Fernandes, F.~Chen, A.~Santamaria, T.~Teng, T.~Frioni, D.~Caldwell, S.~Poni, C.~Semini, M.~Gatti, Using deep learning for pruning region detection and plant organ segmentation in dormant spur-pruned grapevines, Precision Agriculture 24~(4) (2023) 1547--1569.

\bibitem{chen2023hob}
Z.~Chen, K.~Granland, R.~Newbury, C.~Chen, Hob-cnn: Hallucination of occluded branches with a convolutional neural network for 2d fruit trees, Smart Agricultural Technology 3 (2023) 100096.

\bibitem{kok2023obscured}
E.~Kok, X.~Wang, C.~Chen, Obscured tree branches segmentation and 3d reconstruction using deep learning and geometrical constraints, Computers and electronics in agriculture 210 (2023) 107884.

\bibitem{tong2023image}
S.~Tong, J.~Zhang, W.~Li, Y.~Wang, F.~Kang, An image-based system for locating pruning points in apple trees using instance segmentation and rgb-d images, Biosystems Engineering 236 (2023) 277--286.

\bibitem{gentilhomme2023towards}
T.~Gentilhomme, M.~Villamizar, J.~Corre, J.-M. Odobez, \href{https://www.sciencedirect.com/science/article/pii/S0168169923001242}{Towards smart pruning: Vinet, a deep-learning approach for grapevine structure estimation}, Computers and Electronics in Agriculture 207 (2023) 107736.
\newblock \href {https://doi.org/https://doi.org/10.1016/j.compag.2023.107736} {\path{doi:https://doi.org/10.1016/j.compag.2023.107736}}.
\newline\urlprefix\url{https://www.sciencedirect.com/science/article/pii/S0168169923001242}

\bibitem{chen_grapevine_2024}
Z.~Chen, Y.~Wang, S.~Tong, C.~Chen, F.~Kang, \href{https://www.mdpi.com/2076-3417/14/8/3327}{Grapevine {Branch} {Recognition} and {Pruning} {Point} {Localization} {Technology} {Based} on {Image} {Processing}}, Applied Sciences 14~(8) (2024) 3327, number: 8 Publisher: Multidisciplinary Digital Publishing Institute.
\newblock \href {https://doi.org/10.3390/app14083327} {\path{doi:10.3390/app14083327}}.
\newline\urlprefix\url{https://www.mdpi.com/2076-3417/14/8/3327}

\bibitem{girshick2015region}
R.~Girshick, J.~Donahue, T.~Darrell, J.~Malik, Region-based convolutional networks for accurate object detection and segmentation, IEEE transactions on pattern analysis and machine intelligence 38~(1) (2015) 142--158.

\bibitem{girshick2015fast}
R.~Girshick, Fast r-cnn, in: Proceedings of the IEEE international conference on computer vision, 2015, pp. 1440--1448.

\bibitem{ren2015faster}
S.~Ren, K.~He, R.~Girshick, J.~Sun, Faster r-cnn: Towards real-time object detection with region proposal networks, Advances in neural information processing systems 28 (2015).

\bibitem{mask_rcnn}
K.~He, G.~Gkioxari, P.~Dollár, R.~Girshick, Mask r-cnn, in: 2017 IEEE International Conference on Computer Vision (ICCV), 2017, pp. 2980--2988.
\newblock \href {https://doi.org/10.1109/ICCV.2017.322} {\path{doi:10.1109/ICCV.2017.322}}.

\bibitem{cai2019cascade}
Z.~Cai, N.~Vasconcelos, Cascade r-cnn: High quality object detection and instance segmentation, IEEE transactions on pattern analysis and machine intelligence 43~(5) (2019) 1483--1498.

\bibitem{liu2021swin}
Z.~Liu, Y.~Lin, Y.~Cao, H.~Hu, Y.~Wei, Z.~Zhang, S.~Lin, B.~Guo, Swin transformer: Hierarchical vision transformer using shifted windows, in: Proceedings of the IEEE/CVF international conference on computer vision, 2021, pp. 10012--10022.

\bibitem{cheng2020hierarchical}
X.~Cheng, Y.~Zhong, M.~Harandi, Y.~Dai, X.~Chang, H.~Li, T.~Drummond, Z.~Ge, Hierarchical neural architecture search for deep stereo matching, Advances in neural information processing systems 33 (2020) 22158--22169.

\bibitem{robbeberger_unet_2015}
O.~Ronneberger, P.~Fischer, T.~Brox, U-net: Convolutional networks for biomedical image segmentation, in: N.~Navab, J.~Hornegger, W.~M. Wells, A.~F. Frangi (Eds.), Medical Image Computing and Computer-Assisted Intervention -- MICCAI 2015, Springer International Publishing, Cham, 2015, pp. 234--241.

\bibitem{graham1972efficient}
R.~L. Graham, An efficient algorithm for determining the convex hull of a finite planar set, Info. Proc. Lett. 1 (1972) 132--133.

\bibitem{krizhevsky2012imagenet}
A.~Krizhevsky, I.~Sutskever, G.~E. Hinton, Imagenet classification with deep convolutional neural networks, Advances in neural information processing systems 25 (2012).

\bibitem{zhou2019unet++}
Z.~Zhou, M.~M.~R. Siddiquee, N.~Tajbakhsh, J.~Liang, Unet++: Redesigning skip connections to exploit multiscale features in image segmentation, IEEE transactions on medical imaging 39~(6) (2019) 1856--1867.

\bibitem{badrinarayanan2017segnet}
V.~Badrinarayanan, A.~Kendall, R.~Cipolla, Segnet: A deep convolutional encoder-decoder architecture for image segmentation, IEEE transactions on pattern analysis and machine intelligence 39~(12) (2017) 2481--2495.

\bibitem{simonyan2014very}
K.~Simonyan, A.~Zisserman, Very deep convolutional networks for large-scale image recognition, arXiv preprint arXiv:1409.1556 (2014).

\bibitem{long2015fully}
J.~Long, E.~Shelhamer, T.~Darrell, Fully convolutional networks for semantic segmentation, in: Proceedings of the IEEE conference on computer vision and pattern recognition, 2015, pp. 3431--3440.

\bibitem{wang2020solov2}
X.~Wang, R.~Zhang, T.~Kong, L.~Li, C.~Shen, Solov2: Dynamic and fast instance segmentation, Advances in Neural information processing systems 33 (2020) 17721--17732.

\bibitem{zhao2017pyramid}
H.~Zhao, J.~Shi, X.~Qi, X.~Wang, J.~Jia, Pyramid scene parsing network, in: Proceedings of the IEEE conference on computer vision and pattern recognition, 2017, pp. 2881--2890.

\bibitem{newell2016stacked}
A.~Newell, K.~Yang, J.~Deng, Stacked hourglass networks for human pose estimation, in: Computer Vision--ECCV 2016: 14th European Conference, Amsterdam, The Netherlands, October 11-14, 2016, Proceedings, Part VIII 14, Springer, 2016, pp. 483--499.

\bibitem{williams2024archie}
H.~Williams, D.~Smith, J.~Shahabi, T.~Gee, A.~Qureshi, B.~McGuinness, S.~Harvey, C.~Downes, R.~Jangali, K.~Black, et~al., Archie jnr: A robotic platform for autonomous cane pruning of grapevines, in: 2024 IEEE/RSJ International Conference on Intelligent Robots and Systems (IROS), IEEE, 2024, pp. 11736--11743.

\bibitem{wu2019detectron2}
Y.~Wu, A.~Kirillov, F.~Massa, W.-Y. Lo, R.~Girshick, Detectron2, \url{https://github.com/facebookresearch/detectron2} (2019).

\bibitem{kirillov2019panoptic}
A.~Kirillov, K.~He, R.~Girshick, C.~Rother, P.~Doll{\'a}r, Panoptic segmentation, in: Proceedings of the IEEE/CVF conference on computer vision and pattern recognition, 2019, pp. 9404--9413.

\bibitem{ilg2017flownet}
E.~Ilg, N.~Mayer, T.~Saikia, M.~Keuper, A.~Dosovitskiy, T.~Brox, Flownet 2.0: Evolution of optical flow estimation with deep networks, in: Proceedings of the IEEE conference on computer vision and pattern recognition, 2017, pp. 2462--2470.

\bibitem{isola2017image}
P.~Isola, J.-Y. Zhu, T.~Zhou, A.~A. Efros, Image-to-image translation with conditional adversarial networks, in: Proceedings of the IEEE conference on computer vision and pattern recognition, 2017, pp. 1125--1134.

\bibitem{landrieu2018large}
L.~Landrieu, M.~Simonovsky, Large-scale point cloud semantic segmentation with superpoint graphs, in: Proceedings of the IEEE conference on computer vision and pattern recognition, 2018, pp. 4558--4567.

\bibitem{elfiky2015automation}
N.~M. Elfiky, S.~A. Akbar, J.~Sun, J.~Park, A.~Kak, Automation of dormant pruning in specialty crop production: An adaptive framework for automatic reconstruction and modeling of apple trees, in: Proceedings of the IEEE conference on computer vision and pattern recognition workshops, 2015, pp. 65--73.

\bibitem{chattopadhyay2016measuring}
S.~Chattopadhyay, S.~A. Akbar, N.~M. Elfiky, H.~Medeiros, A.~Kak, Measuring and modeling apple trees using time-of-flight data for automation of dormant pruning applications, in: 2016 IEEE Winter conference on applications of computer vision (WACV), IEEE, 2016, pp. 1--9.

\bibitem{medeiros_modeling_2017}
H.~Medeiros, D.~Kim, J.~Sun, H.~Seshadri, S.~A. Akbar, N.~M. Elfiky, J.~Park, \href{https://onlinelibrary.wiley.com/doi/10.1002/rob.21679}{Modeling {Dormant} {Fruit} {Trees} for {Agricultural} {Automation}}, Journal of Field Robotics 34~(7) (2017) 1203--1224.
\newblock \href {https://doi.org/10.1002/rob.21679} {\path{doi:10.1002/rob.21679}}.
\newline\urlprefix\url{https://onlinelibrary.wiley.com/doi/10.1002/rob.21679}

\bibitem{liu_research_2019}
S.~Liu, J.~Yao, H.~Li, C.~Qiu, R.~Liu, \href{https://iopscience.iop.org/article/10.1088/1742-6596/1237/2/022059}{Research on {3D} skeletal model extraction algorithm of branch based on {SR4000}}, Journal of Physics: Conference Series 1237~(2) (2019) 022059.
\newblock \href {https://doi.org/10.1088/1742-6596/1237/2/022059} {\path{doi:10.1088/1742-6596/1237/2/022059}}.
\newline\urlprefix\url{https://iopscience.iop.org/article/10.1088/1742-6596/1237/2/022059}

\bibitem{xu2022improved}
Y.~Xu, C.~Hu, Y.~Xie, An improved space colonization algorithm with dbscan clustering for a single tree skeleton extraction, International Journal of Remote Sensing 43~(10) (2022) 3692--3713.

\bibitem{you2023tree}
L.~You, Y.~Sun, Y.~Liu, X.~Chang, J.~Jiang, Y.~Feng, X.~Song, Tree skeletonization with dbscan clustering using terrestrial laser scanning data, Forests 14~(8) (2023) 1525.

\bibitem{fu2023skeleton}
Y.~Fu, Y.~Xia, H.~Zhang, M.~Fu, Y.~Wang, W.~Fu, C.~Shen, Skeleton extraction and pruning point identification of jujube tree for dormant pruning using space colonization algorithm, Frontiers in Plant Science 13 (2023) 1103794.

\bibitem{li2024efficient}
X.~Li, B.~Liu, Y.~Shi, M.~Xiong, D.~Ren, L.~Wu, X.~Zou, Efficient three-dimensional reconstruction and skeleton extraction for intelligent pruning of fruit trees, Computers and Electronics in Agriculture 227 (2024) 109554.

\bibitem{dukic2024branch}
J.~Duki{\'c}, P.~Peji{\'c}, A.~Bo{\v{s}}njak, E.~K. Nyarko, Branch-a labeled dataset of rgb-d images and 3d models for autonomous tree pruning, in: 2024 International Conference on Smart Systems and Technologies (SST), IEEE, 2024, pp. 57--64.

\bibitem{besl1992method}
P.~J. Besl, N.~D. McKay, Method for registration of 3-d shapes, in: Sensor fusion IV: control paradigms and data structures, Vol. 1611, Spie, 1992, pp. 586--606.

\bibitem{su2022adaptive}
T.~Su, W.~Wang, H.~Liu, Z.~Liu, X.~Li, Z.~Jia, L.~Zhou, Z.~Song, M.~Ding, A.~Cui, An adaptive and rapid 3d delaunay triangulation for randomly distributed point cloud data, The Visual Computer (2022) 1--25.

\bibitem{straub2022approach}
J.~Straub, D.~Reiser, N.~L{\"u}ling, A.~Stana, H.~W. Griepentrog, Approach for graph-based individual branch modelling of meadow orchard trees with 3d point clouds, Precision Agriculture 23~(6) (2022) 1967--1982.

\bibitem{hu2020non}
G.~Hu, Z.~Zhou, J.~Cao, H.~Huang, Non-linear calibration optimisation based on the levenberg--marquardt algorithm, IET Image Processing 14~(7) (2020) 1402--1414.

\bibitem{tabb2017robotic}
A.~Tabb, H.~Medeiros, A robotic vision system to measure tree traits, in: 2017 IEEE/RSJ International Conference on Intelligent Robots and Systems (IROS), IEEE, 2017, pp. 6005--6012.

\bibitem{dong2024improved}
X.~Dong, W.-Y. Kim, Z.~Yu, J.-Y. Oh, R.~Ehsani, K.-H. Lee, Improved voxel-based volume estimation and pruning severity mapping of apple trees during the pruning period, Computers and Electronics in Agriculture 219 (2024) 108834.

\bibitem{zhu2007stochastic}
S.-C. Zhu, D.~Mumford, et~al., A stochastic grammar of images, Foundations and Trends{\textregistered} in Computer Graphics and Vision 2~(4) (2007) 259--362.

\bibitem{botterill2013finding}
T.~Botterill, R.~Green, S.~Mills, Finding a vine's structure by bottom-up parsing of cane edges, in: 2013 28th International Conference on Image and Vision Computing New Zealand (IVCNZ 2013), IEEE, 2013, pp. 112--117.

\bibitem{ramer1972iterative}
U.~Ramer, An iterative procedure for the polygonal approximation of plane curves, Computer graphics and image processing 1~(3) (1972) 244--256.

\bibitem{zhang_fast_1984}
T.~Y. Zhang, C.~Y. Suen, \href{https://doi.org/10.1145/357994.358023}{A fast parallel algorithm for thinning digital patterns}, Commun. ACM 27~(3) (1984) 236–239.
\newblock \href {https://doi.org/10.1145/357994.358023} {\path{doi:10.1145/357994.358023}}.
\newline\urlprefix\url{https://doi.org/10.1145/357994.358023}

\bibitem{cuevas_segmentation_2020}
H.~Cuevas-Velasquez, A.-J. Gallego, R.~B. Fisher, \href{https://www.sciencedirect.com/science/article/pii/S0168169919323919}{Segmentation and 3d reconstruction of rose plants from stereoscopic images}, Computers and Electronics in Agriculture 171 (2020) 105296.
\newblock \href {https://doi.org/https://doi.org/10.1016/j.compag.2020.105296} {\path{doi:https://doi.org/10.1016/j.compag.2020.105296}}.
\newline\urlprefix\url{https://www.sciencedirect.com/science/article/pii/S0168169919323919}

\bibitem{hilitch1969linear}
C.~J. Hilitch, Linear skeletons from square cupboards (1969).

\bibitem{martin2000image}
A.~Martin, S.~Tosunoglu, Image processing techniques for machine vision, Miami, Florida (2000) 1--9.

\bibitem{cao2010point}
J.~Cao, A.~Tagliasacchi, M.~Olson, H.~Zhang, Z.~Su, Point cloud skeletons via laplacian based contraction, in: 2010 Shape Modeling International Conference, 2010, pp. 187--197.
\newblock \href {https://doi.org/10.1109/SMI.2010.25} {\path{doi:10.1109/SMI.2010.25}}.

\bibitem{harris1988combined}
C.~Harris, M.~Stephens, et~al., A combined corner and edge detector, in: Alvey vision conference, Vol.~15, Citeseer, 1988, pp. 10--5244.

\bibitem{fu2020three}
Y.~Fu, C.~Li, J.~Zhu, B.~Wang, B.~Zhang, W.~Fu, Three-dimensional model construction method and experiment of jujube tree point cloud using alpha-shape algorithm, Trans. Chin. Soc Agric. Eng 36~(22) (2020) 214--221.

\bibitem{runions2005modeling}
A.~Runions, M.~Fuhrer, B.~Lane, P.~Federl, A.-G. Rolland-Lagan, P.~Prusinkiewicz, Modeling and visualization of leaf venation patterns, in: ACM SIGGRAPH 2005 Papers, 2005, pp. 702--711.

\bibitem{akbar2016novel}
S.~A. Akbar, S.~Chattopadhyay, N.~M. Elfiky, A.~Kak, A novel benchmark rgbd dataset for dormant apple trees and its application to automatic pruning, in: Proceedings of the IEEE conference on computer vision and pattern recognition workshops, 2016, pp. 81--88.

\bibitem{marquardt1963algorithm}
D.~W. Marquardt, An algorithm for least-squares estimation of nonlinear parameters, Journal of the society for Industrial and Applied Mathematics 11~(2) (1963) 431--441.

\bibitem{schnabel2007efficient}
R.~Schnabel, R.~Wahl, R.~Klein, Efficient ransac for point-cloud shape detection, in: Computer graphics forum, Vol.~26, Wiley Online Library, 2007, pp. 214--226.

\bibitem{qi2017pointnet++}
C.~R. Qi, L.~Yi, H.~Su, L.~J. Guibas, Pointnet++: Deep hierarchical feature learning on point sets in a metric space, Advances in neural information processing systems 30 (2017).

\bibitem{lin2021point2skeleton}
C.~Lin, C.~Li, Y.~Liu, N.~Chen, Y.-K. Choi, W.~Wang, Point2skeleton: Learning skeletal representations from point clouds, in: Proceedings of the IEEE/CVF conference on computer vision and pattern recognition, 2021, pp. 4277--4286.

\bibitem{yandun2020visual}
F.~Yandun, A.~Silwal, G.~Kantor, Visual 3d reconstruction and dynamic simulation of fruit trees for robotic manipulation, in: Proceedings of the IEEE/CVF Conference on Computer Vision and Pattern Recognition Workshops, 2020, pp. 54--55.

\bibitem{guadagna202116}
P.~Guadagna, T.~Frioni, F.~Chen, A.~I. Delmonte, T.~Teng, M.~Fernandes, A.~Scaldaferri, C.~Semini, S.~Poni, M.~Gatti, 16. fine-tuning and testing of a deep learning algorithm for pruning regions detection in spur-pruned grapevines, in: Precision agriculture'21, Wageningen Academic, 2021, pp. 147--153.

\bibitem{silwal2022bumblebee}
A.~Silwal, F.~Yandun, A.~K. Nellithimaru, T.~Bates, G.~Kantor, Bumblebee: A path towards fully autonomous robotic vine pruning., Field Robotics 2~(1) (2022) 1661--1696.

\bibitem{you_autonomous_2022}
A.~You, N.~Parayil, J.~G. Krishna, U.~Bhattarai, R.~Sapkota, D.~Ahmed, M.~Whiting, M.~Karkee, C.~M. Grimm, J.~R. Davidson, \href{http://arxiv.org/abs/2206.07201}{An autonomous robot for pruning modern, planar fruit trees}, arXiv:2206.07201 [cs] (Jun. 2022).
\newline\urlprefix\url{http://arxiv.org/abs/2206.07201}

\bibitem{strnad2017novel}
D.~Strnad, {\v{S}}.~Kohek, Novel discrete differential evolution methods for virtual tree pruning optimization, Soft Computing 21~(4) (2017) 981--993.

\bibitem{kolmanivc2021algorithm}
S.~Kolmani{\v{c}}, D.~Strnad, {\v{S}}.~Kohek, B.~Benes, P.~Hirst, B.~{\v{Z}}alik, An algorithm for automatic dormant tree pruning, Applied Soft Computing 99 (2021) 106931.

\bibitem{marset2021towards}
W.~V. Marset, D.~S. P{\'e}rez, C.~A. D{\'\i}az, F.~Bromberg, Towards practical 2d grapevine bud detection with fully convolutional networks, Computers and Electronics in Agriculture 182 (2021) 105947.

\bibitem{diaz2018grapevine}
C.~A. D{\'\i}az, D.~S. P{\'e}rez, H.~Miatello, F.~Bromberg, Grapevine buds detection and localization in 3d space based on structure from motion and 2d image classification, Computers in Industry 99 (2018) 303--312.

\bibitem{zhao_multiple_2024}
G.~Zhao, D.~Wang, \href{https://www.mdpi.com/2624-7402/6/1/33}{A {Multiple} {Criteria} {Decision}-{Making} {Method} {Generated} by the {Space} {Colonization} {Algorithm} for {Automated} {Pruning} {Strategies} of {Trees}}, AgriEngineering 6~(1) (2024) 539--554.
\newblock \href {https://doi.org/10.3390/agriengineering6010033} {\path{doi:10.3390/agriengineering6010033}}.
\newline\urlprefix\url{https://www.mdpi.com/2624-7402/6/1/33}

\bibitem{tang2015integrated}
L.~Tang, C.~Chen, H.~Huang, D.~Lin, An integrated system for 3d tree modeling and growth simulation, Environmental Earth Sciences 74 (2015) 7015--7028.

\bibitem{berut2018gravisensors}
A.~B{\'e}rut, H.~Chauvet, V.~Legu{\'e}, B.~Moulia, O.~Pouliquen, Y.~Forterre, Gravisensors in plant cells behave like an active granular liquid, Proceedings of the National Academy of Sciences 115~(20) (2018) 5123--5128.

\bibitem{mvech1996visual}
R.~M{\v{e}}ch, P.~Prusinkiewicz, Visual models of plants interacting with their environment, in: Proceedings of the 23rd annual conference on Computer graphics and interactive techniques, 1996, pp. 397--410.

\bibitem{palubicki2009self}
W.~Palubicki, K.~Horel, S.~Longay, A.~Runions, B.~Lane, R.~M{\v{e}}ch, P.~Prusinkiewicz, Self-organizing tree models for image synthesis, ACM Transactions On Graphics (TOG) 28~(3) (2009) 1--10.

\bibitem{corbett-davies_expert_2012}
S.~Corbett-Davies, T.~Botterill, R.~Green, V.~Saxton, \href{https://dl.acm.org/doi/10.1145/2425836.2425849}{An expert system for automatically pruning vines}, in: Proceedings of the 27th {Conference} on {Image} and {Vision} {Computing} {New} {Zealand}, ACM, Dunedin New Zealand, 2012, pp. 55--60.
\newblock \href {https://doi.org/10.1145/2425836.2425849} {\path{doi:10.1145/2425836.2425849}}.
\newline\urlprefix\url{https://dl.acm.org/doi/10.1145/2425836.2425849}

\bibitem{runions2007modeling}
A.~Runions, B.~Lane, P.~Prusinkiewicz, Modeling trees with a space colonization algorithm., Nph 7~(63-70) (2007) 6.

\bibitem{bryson2023using}
M.~Bryson, F.~Wang, J.~Allworth, Using synthetic tree data in deep learning-based tree segmentation using lidar point clouds, Remote Sensing 15~(9) (2023) 2380.

\bibitem{kohek2015eduapple}
{\v{S}}.~Kohek, N.~Guid, S.~Tojnko, T.~Unuk, S.~Kolmani{\v{c}}, Eduapple: Interactive teaching tool for apple tree crown formation, HortTechnology 25~(2) (2015) 238--246.

\bibitem{kang2016imapple}
H.~Kang, M.~Fiser, B.~Shi, F.~Sheibani, P.~Hirst, B.~Benes, Imapple—functional structural model of apple trees, in: 2016 IEEE international conference on functional-structural plant growth modeling, simulation, visualization and applications (FSPMA), IEEE, 2016, pp. 90--97.

\bibitem{de1988plant}
P.~De~Reffye, C.~Edelin, J.~Fran{\c{c}}on, M.~Jaeger, C.~Puech, Plant models faithful to botanical structure and development, ACM Siggraph Computer Graphics 22~(4) (1988) 151--158.

\bibitem{storn1997differential}
R.~Storn, K.~Price, Differential evolution--a simple and efficient heuristic for global optimization over continuous spaces, Journal of global optimization 11 (1997) 341--359.

\bibitem{westling_procedure_2021}
F.~Westling, J.~Underwood, M.~Bryson, \href{http://arxiv.org/abs/2102.03700}{A procedure for automated tree pruning suggestion using {LiDAR} scans of fruit trees}, arXiv:2102.03700 [cs, eess] (Feb. 2021).
\newline\urlprefix\url{http://arxiv.org/abs/2102.03700}

\bibitem{hart1968formal}
P.~E. Hart, N.~J. Nilsson, B.~Raphael, A formal basis for the heuristic determination of minimum cost paths, IEEE transactions on Systems Science and Cybernetics 4~(2) (1968) 100--107.

\bibitem{qiu20243d}
T.~Qiu, A.~Zoubi, L.~Cheng, Y.~Jiang, 3d branch point cloud completion for robotic pruning in apple orchards, arXiv preprint arXiv:2404.05953 (2024).

\bibitem{qiu20223d_characterization}
T.~Qiu, L.~Cheng, Y.~Jiang, 3d characterization of tree architecture for apple crop load estimation, in: 2022 ASABE Annual International Meeting, American Society of Agricultural and Biological Engineers, 2022, p.~1.

\bibitem{yu2023adapointr}
X.~Yu, Y.~Rao, Z.~Wang, J.~Lu, J.~Zhou, Adapointr: Diverse point cloud completion with adaptive geometry-aware transformers, IEEE Transactions on Pattern Analysis and Machine Intelligence (2023).

\bibitem{yang2017obstacle}
H.~Yang, L.~Li, Z.~Gao, Obstacle avoidance path planning of hybrid harvesting manipulator based on joint configuration space, Transactions of the Chinese Society of Agricultural Engineering 33~(4) (2017) 55--62.

\bibitem{bac2017performance}
C.~W. Bac, J.~Hemming, B.~Van~Tuijl, R.~Barth, E.~Wais, E.~J. van Henten, Performance evaluation of a harvesting robot for sweet pepper, Journal of Field Robotics 34~(6) (2017) 1123--1139.

\bibitem{harrell1990robotic}
R.~Harrell, P.~D. Adsit, R.~Munilla, D.~Slaughter, Robotic picking of citrus, Robotica 8~(4) (1990) 269--278.

\bibitem{silwal2017design}
A.~Silwal, J.~R. Davidson, M.~Karkee, C.~Mo, Q.~Zhang, K.~Lewis, Design, integration, and field evaluation of a robotic apple harvester, Journal of Field Robotics 34~(6) (2017) 1140--1159.

\bibitem{tanigaki2008cherry}
K.~Tanigaki, T.~Fujiura, A.~Akase, J.~Imagawa, Cherry-harvesting robot, Computers and electronics in agriculture 63~(1) (2008) 65--72.

\bibitem{zhang2015kinematics}
J.~Zhang, J.~K. Schueller, Kinematics and dynamics of a fruit picking robotic manipulator, in: 2015 ASABE Annual International Meeting, American Society of Agricultural and Biological Engineers, 2015, p.~1.

\bibitem{li2023automatic}
Y.~Li, Z.~Zhang, X.~Wang, W.~Fu, J.~Li, Automatic reconstruction and modeling of dormant jujube trees using three-view image constraints for intelligent pruning applications, Computers and Electronics in Agriculture 212 (2023) 108149.

\bibitem{you_efficient_2020}
A.~You, F.~Sukkar, R.~Fitch, M.~Karkee, J.~R. Davidson, An {Efficient} {Planning} and {Control} {Framework} for {Pruning} {Fruit} {Trees}, in: 2020 {IEEE} {International} {Conference} on {Robotics} and {Automation} ({ICRA}), 2020, pp. 3930--3936, iSSN: 2577-087X.
\newblock \href {https://doi.org/10.1109/ICRA40945.2020.9197551} {\path{doi:10.1109/ICRA40945.2020.9197551}}.

\bibitem{teng_whole-body_2021}
T.~Teng, M.~Fernandes, M.~Gatti, S.~Poni, C.~Semini, D.~Caldwell, F.~Chen, Whole-{Body} {Control} on {Non}-holonomic {Mobile} {Manipulation} for {Grapevine} {Winter} {Pruning} {Automation}, in: 2021 6th {IEEE} {International} {Conference} on {Advanced} {Robotics} and {Mechatronics} ({ICARM}), 2021, pp. 37--42.
\newblock \href {https://doi.org/10.1109/ICARM52023.2021.9536083} {\path{doi:10.1109/ICARM52023.2021.9536083}}.

\bibitem{zhang2022design}
B.~Zhang, X.~Chen, H.~Zhang, C.~Shen, W.~Fu, Design and performance test of a jujube pruning manipulator, Agriculture 12~(4) (2022) 552.

\bibitem{zahid2020development}
A.~Zahid, L.~He, L.~Zeng, D.~Choi, J.~Schupp, P.~Heinemann, Development of a robotic end-effector for apple tree pruning, Transactions of the ASABE 63~(4) (2020) 847--856.

\bibitem{lu_kinematics_2021}
Y.~Lu, \href{https://iopscience.iop.org/article/10.1088/1755-1315/769/4/042067}{Kinematics {Analysis} and {Trajectory} {Planning} of {Dual}-arm {Pruning} {Robot}}, IOP Conference Series: Earth and Environmental Science 769~(4) (2021) 042067.
\newblock \href {https://doi.org/10.1088/1755-1315/769/4/042067} {\path{doi:10.1088/1755-1315/769/4/042067}}.
\newline\urlprefix\url{https://iopscience.iop.org/article/10.1088/1755-1315/769/4/042067}

\bibitem{kondo1998robotics}
N.~Kondo, K.~Ting*, Robotics for plant production, Artificial intelligence review 12 (1998) 227--243.

\bibitem{you2023semiautonomous}
A.~You, N.~Parayil, J.~G. Krishna, U.~Bhattarai, R.~Sapkota, D.~Ahmed, M.~Whiting, M.~Karkee, C.~M. Grimm, J.~R. Davidson, Semiautonomous precision pruning of upright fruiting offshoot orchard systems: An integrated approach, IEEE Robotics \& Automation Magazine (2023).

\bibitem{slotine1991general}
S.~B. Slotine, B.~Siciliano, A general framework for managing multiple tasks in highly redundant robotic systems, in: proceeding of 5th International Conference on Advanced Robotics, Vol.~2, 1991, pp. 1211--1216.

\bibitem{guan2005robotic}
Y.~Guan, K.~Yokoi, O.~Stasse, A.~Kheddar, On robotic trajectory planning using polynomial interpolations, in: 2005 IEEE international conference on robotics and biomimetics-ROBIO, IEEE, 2005, pp. 111--116.

\bibitem{kuffner2000rrt}
J.~J. Kuffner, S.~M. LaValle, Rrt-connect: An efficient approach to single-query path planning, in: Proceedings 2000 ICRA. Millennium Conference. IEEE International Conference on Robotics and Automation. Symposia Proceedings (Cat. No. 00CH37065), Vol.~2, IEEE, 2000, pp. 995--1001.

\bibitem{chen2022path}
Y.~Chen, Y.~Fu, B.~Zhang, W.~Fu, C.~Shen, Path planning of the fruit tree pruning manipulator based on improved rrt-connect algorithm, International Journal of Agricultural and Biological Engineering 15~(2) (2022) 177--188.

\bibitem{paulin2015comparison}
S.~Paulin, T.~Botterill, J.~Lin, X.~Chen, R.~Green, A comparison of sampling-based path planners for a grape vine pruning robot arm, in: 2015 6th International Conference on Automation, Robotics and Applications (ICARA), IEEE, 2015, pp. 98--103.

\bibitem{zucker2013chomp}
M.~Zucker, N.~Ratliff, A.~D. Dragan, M.~Pivtoraiko, M.~Klingensmith, C.~M. Dellin, J.~A. Bagnell, S.~S. Srinivasa, Chomp: Covariant hamiltonian optimization for motion planning, The International journal of robotics research 32~(9-10) (2013) 1164--1193.

\bibitem{schulman2013finding}
J.~Schulman, J.~Ho, A.~X. Lee, I.~Awwal, H.~Bradlow, P.~Abbeel, Finding locally optimal, collision-free trajectories with sequential convex optimization., in: Robotics: science and systems, Vol.~9, Berlin, Germany, 2013, pp. 1--10.

\bibitem{gammell2015batch}
J.~D. Gammell, S.~S. Srinivasa, T.~D. Barfoot, Batch informed trees (bit*): Sampling-based optimal planning via the heuristically guided search of implicit random geometric graphs, in: 2015 IEEE international conference on robotics and automation (ICRA), IEEE, 2015, pp. 3067--3074.

\bibitem{silwal2021robust}
A.~Silwal, T.~Parhar, F.~Yandun, H.~Baweja, G.~Kantor, A robust illumination-invariant camera system for agricultural applications, in: 2021 IEEE/RSJ International Conference on Intelligent Robots and Systems (IROS), IEEE, 2021, pp. 3292--3298.

\bibitem{goodfellow2014generative}
I.~Goodfellow, J.~Pouget-Abadie, M.~Mirza, B.~Xu, D.~Warde-Farley, S.~Ozair, A.~Courville, Y.~Bengio, Generative adversarial nets, Advances in neural information processing systems 27 (2014).

\bibitem{james2019sim}
S.~James, P.~Wohlhart, M.~Kalakrishnan, D.~Kalashnikov, A.~Irpan, J.~Ibarz, S.~Levine, R.~Hadsell, K.~Bousmalis, Sim-to-real via sim-to-sim: Data-efficient robotic grasping via randomized-to-canonical adaptation networks, in: Proceedings of the IEEE/CVF conference on computer vision and pattern recognition, 2019, pp. 12627--12637.

\bibitem{lauretti2023robot}
C.~Lauretti, C.~Tamantini, H.~Tom{\`e}, L.~Zollo, Robot learning by demonstration with dynamic parameterization of the orientation: an application to agricultural activities, Robotics 12~(6) (2023) 166.

\bibitem{haring2024vid2cuts}
S.~H{\"a}ring, S.~Folawiyo, M.~Podguzova, D.~Stricker, et~al., Vid2cuts: A framework for enabling ai-guided grapevine pruning, IEEE Access 12 (2024) 5814--5836.

\bibitem{kim2025autonomous}
C.~H. Kim, A.~Silwal, G.~Kantor, Autonomous robotic pepper harvesting: Imitation learning in unstructured agricultural environments, IEEE Robotics and Automation Letters (2025).

\end{thebibliography}

\end{document}